%% file: elsarticle-template-num-clean.tex
\documentclass[preprint,12pt,]{elsarticle}

\usepackage{amssymb}
\usepackage{xcolor}
\usepackage{lscape}
\usepackage{booktabs}
\usepackage{multirow}
\usepackage{nameref}
\usepackage{amsmath}
\usepackage{hyperref}
\usepackage{rotating}
\usepackage{cleveref}
\usepackage{array}

\journal{Energy and AI}

\begin{document}

\begin{frontmatter}



\title{Exploring Public Attention in the Circular Economy through Topic Modelling with Twin Hyperparameter Optimisation}

\author{Junhao Song\fnref{label1,label2}\corref{cor1}}
\ead{junhao.song23@imperial.ac.uk}

\author{Yingfang Yuan\fnref{label1}\corref{cor1}}
\ead{y.yuan@hw.ac.uk}

\author{Kaiwen Chang\fnref{label5}}
\ead{003867@nuist.edu.cn}

\author{Bing Xu\fnref{label3}}
\ead{b.xu@hw.ac.uk}

\author{Jin Xuan\fnref{label4}}
\ead{j.xuan@surrey.ac.uk}

\author{Wei Pang\corref{cor2}\fnref{label1}}
\ead{w.pang@hw.ac.uk}
\cortext[cor1]{Both authors contributed equally to this research.}
\cortext[cor2]{Corresponding author.}

\affiliation[label1]{organization={School of Mathematical and Computer Sciences, Heriot-Watt University},
            city={Edinburgh},
            country={United Kingdom}}

\affiliation[label2]{organization={Faculty of Engineering, Imperial College London},
            city={London},
            country={United Kingdom}}

\affiliation[label3]{organization={Edinburgh Business School, Heriot-Watt University},
            city={Edinburgh},
            country={United Kingdom}}

\affiliation[label4]{organization={Faculty of Engineering and Physical Sciences, University of Surrey},
            city={Surrey},
            country={United Kingdom}}

\affiliation[label5]{organization={School of Management Science and Engineering, Nanjing University of Information Science and Technology},
            city={Nanjing},
            country={China}}

\begin{abstract}
To advance the circular economy (CE), it is crucial to gain insights into the evolution of public attention, cognitive pathways of the masses concerning circular products, and to identify primary concerns. To achieve this, we collected data from diverse platforms, including Twitter, Reddit, and The Guardian, and utilised three topic models to analyse the data. Given the performance of topic modelling may vary depending on hyperparameter settings, this research proposed a novel framework that integrates twin (single and multi-objective) hyperparameter optimisation for the CE. We conducted systematic experiments to ensure that topic models are set with appropriate hyperparameters under different constraints, providing valuable insights into the correlations between CE and public attention. In summary, our optimised model reveals that public remains concerned about the economic impacts of sustainability and circular practices, particularly regarding recyclable materials and environmentally sustainable technologies. The analysis shows that the CE has attracted significant attention on The Guardian, especially in topics related to sustainable development and environmental protection technologies, while discussions are comparatively less active on Twitter. These insights highlight the need for policymakers to implement targeted education programs, create incentives for businesses to adopt CE principles, and enforce more stringent waste management policies alongside improved recycling processes.
\end{abstract}



\end{frontmatter}



\input{section1_clean}

\input{section2_clean}
\input{section3_clean}
\input{section4_clean}

\input{section5_clean}
\appendix
\section{}

\begin{table*}[ht]
\centering
\caption{53 keywords related to circular economy for data scraping. Based on the existing literature, this study first summarises the keywords related to the circular economy \cite{kirchherr2017conceptualizing, arruda2021, Chauhan2022}. Then, these keywords are expanded using the word embedding model to develop a comprehensive circular economy dictionary. As a novel variable quantification method, the word embedding model improves the accuracy of variable measurement and enhances the robustness of empirical research results \cite{li2021measuring}. This model employs neural networks to perform deep analysis on a large volume of economic and financial texts, generating a word similarity model that can identify and train similar words. The similarity dictionary produced by this model allows variables to be measured more comprehensively and objectively \cite{li2021measuring}.}
\footnotesize
\begin{tabular}{|>{\raggedright\arraybackslash}p{3.5cm}|>{\raggedright\arraybackslash}p{3.5cm}|>{\raggedright\arraybackslash}p{3.5cm}|}
\hline
sustainable development & energy saving and environmental protection & energy saving \\
\hline
green economy & eco-design & energy conservation \\
\hline
clean manufacturing & reuse & zero waste \\
\hline
renewable energy & reducing carbon footprint & renewable resources \\
\hline
waste reduction & low carbon & closed-loop system \\
\hline
energy efficiency & green manufacturing & environment friendly \\
\hline
green procurement & recycling & environmentally friendly \\
\hline
green & natural capital & biomimicry \\
\hline
social responsibility & environmental economics & efficient use \\
\hline
regeneration cycle & sustainable supply chain & industrial ecology \\
\hline
durability & environmental impact assessment & resource efficiency \\
\hline
sharing economy & eco-friendly products & save resources \\
\hline
green low carbon & clean production & sustainable consumption \\
\hline
eco-efficiency & resource saving & clean \\
\hline
environmental protection & clean energy & biodegradable material \\
\hline
extended producer responsibility & ecological civilization & environmental governance \\
\hline
zero emission & resource recovery & remanufacturing \\
\hline
waste management & green ecology & \\
\hline
\end{tabular}
\label{appendix:keywords}
\end{table*}

\bibliographystyle{elsarticle-num-clean} 



\bibliography{mybib_clean}

\end{document}

%% file: section1_clean.tex
\section{Introduction}

The Circular Economy (CE) has gained increased attention from both academics and practitioners in recent years. The primary objectives of the CE are commonly regarded as achieving the twin goals of economic prosperity and improved resource efficiency \cite{kirchherr2017conceptualizing}. However, its impacts on social equity and public perceptions are often overlooked. CE also plays a critical role in transitioning towards net-zero emissions \cite{corvellec2022critiques} by reducing resource consumption and optimising waste management \cite{morseletto2020}, which lowers energy demand and greenhouse gas emissions \cite{oenema2001technical}. This relationship is especially significant in fields such as energy and artificial intelligence, where integrating advanced technologies and circular principles can lead to substantial improvements in resource efficiency and sustainability.

Moreover, public attention and understanding of CE remain limited. CE represents a paradigm shift from the traditional linear economy—characterised by the `take–make–waste' model \cite{macarthur2015towards} to one that promotes the principles of regeneration, material reuse, waste reduction, and pollution prevention. In a world grappling with pressing environmental challenges, public discourse on CE plays a pivotal role in shaping policies \cite{arruda2021}, business strategies \cite{velenturf2021}, and societal perceptions of sustainable development \cite{bataille2020phy}. Therefore, it has become increasingly crucial to comprehend and interpret public attention concerning the CE. This urgency stems not only from the necessity to understand the intricacies of evolving public attitudes but also from the imperative to analyse the cognitive pathways that influence the public's perception of recycled products within the broader framework of the CE \cite{korhonen2018}.

Throughout the process of our research, we meticulously designed and executed a series of systematic experiments with the primary goal of unravelling the intricate relationship between CE and public attention. We aimed to move beyond mere observation and dive into the underlying factors driving changes in attention. To achieve this, we employed topic modelling techniques \cite{wallach2009evaluation}. Topic modelling techniques are a class of machine learning algorithms used to automatically discover the underlying themes or `topics' present in large collections of unstructured text data. 

To be more specific, we utilised three well-established and open-source topic modelling techniques for their reliability and effectiveness. These include models based on Latent Dirichlet Allocation (LDA) technology \cite{blei2003latent}, Correlation Explanation (CorEx) \cite{gallagher2018corex}, and BERTopic model \cite{grootendorst2022}. Using LDA, CorEx, and BERTopic together can provide a more comprehensive understanding of the underlying topics in a dataset, as each technique offers distinct strengths. LDA’s probabilistic model helps identify clear, interpretable topics based on word co-occurrence patterns. CorEx, with its focus on maximising correlations, can discover more nuanced relationships that might not be captured by word frequency alone, leading to more coherent and interpretable topics, especially in complex or highly correlated datasets. BERTopic, which leverages BERT embedding, adds the ability to capture contextual and semantic nuances between words, which is often missed by traditional models like LDA. In this research, we selected three datasets: Twitter, The Guardian, and Reddit for model training. As an old news media, The Guardian has rich official information. In particular, it focuses on the attention of organisations such as governments on CE. Social media platforms such as Twitter and Reddit are rich sources of real-time user-generated content, where people can express their views and ideas on CE. By analysing the data from these platforms, we can capture diverse and contemporary public attention, and thus understand the nuances of public attention to CE around the world.

Furthermore, this research delves into the complexity of the three selected models. We not only trained different models across various datasets but also fine-tuned the hyperparameters of each model using our novel \textbf{Twin} method (both single-objective and multi-objective optimisation techniques). Hyperparameter optimisation is crucial for topic modelling because the performance and quality of the topics generated by models like LDA, CorEx, and BERTopic are sensitive to the selection of hyperparameters. Hyperparameters such as the number of topics can greatly influence the interpretability, coherence, and relevance of the topics. By tuning these hyperparameters, we can ensure that the model effectively captures the underlying structure of the data and produces meaningful, interpretable topics. Without proper optimisation, the model risks either underfitting or overfitting, resulting in irrelevant or overly broad topics. Hyperparameter optimisation explores the hyperparameter space to find optimal configurations, ensuring the best possible performance and more reliable results.

The hyperparameter optimisation processes employed grid search and the Tree-structured Parzen Estimator (TPE), respectively. Optimal hyperparameters were identified for different datasets based on \textbf{C}oherence using \textbf{N}ormalised \textbf{P}ointwise \textbf{M}utual \textbf{I}nformation ($C_{\text{NPMI}}$) scores \cite{church1990, bouma2009NormalizedM}, as well as diversity \cite{roder2015exploring} and perplexity \cite{wallach2009evaluation}. This approach ensures the efficacy and robustness of our analysis from multiple criteria. Through the hyperparameter optimisation of the three models, this research found BERTopic performed best on both The Guardian and Reddit datasets. CorEx achieved the highest $C_{\text{NPMI}}$ score on the Twitter dataset.

This research offers several key contributions.  First, we collected data from a wide range of sources, including posts from widely used social media platforms (Twitter and Reddit), capturing a wide range of public views on the CE. By analysing the content from The Guardian further reveals distinct differences in how platforms disseminate information and engage in CE discussions. Furthermore, our findings provide a comprehensive view of evolving public perceptions of CE, capturing shifts in attention and attitudes across diverse societal segments.  In essence, this research pioneers the exploration of public attention across different facets of the CE, offering valuable insights into the pathways that shape policies, corporate strategies, and societal attitudes towards a sustainable future. This work contributes to the global effort to foster a more sustainable CE.

The rest of the paper is organised as follows. In sections \ref{sec:relatedwork} and \ref{sec:methodology}, we will discuss the related work, the proposed framework for this research, and twin hyperparameter optimisation method, as well as the experimental settings, respectively. In Section \ref{sec:experiments}, we will comprehensively discuss and analyse each optimal model's generated topics related to the CE, as well as the evolution of public attention towards CE topics over the past few years. Section 5 offers concluding remarks, policy implications and pinpoints potential avenues for future studies.

%% file: section2_clean.tex
\section{Related work}\label{sec:relatedwork}

In this section, we thoroughly examine the extant research and approaches relevant to our research's scope and objectives. Our research is primarily oriented towards discerning public concerns regarding pivotal aspects of the CE by employing advanced techniques such as topic modelling and the utilisation of language models. To facilitate a structured and comprehensive review, we categorise the related work into six subsections, each dedicated to an essential facet of our research: \nameref{subsec:lda}, \nameref{subsec:corex}, \nameref{subsec:bertopic}, \nameref{subsec:metrics}, and \nameref{subsec:circular}. This section is intended to provide readers with a profound understanding of the foundational research that underpins our investigation.

\subsection{Latent Dirichlet Allocation (LDA)}\label{subsec:lda}

LDA is a technique for discovering the underlying topic structure inherent in a corpus of text documents. LDA operates within the framework of unsupervised machine learning \cite{barlow1989} and provides a way to convert inherently qualitative textual information into a quantitative format that can be used for rigorous statistical modelling by representing each document as a vector of word frequencies. LDA assumes that each document in a collection of documents can be conceptually understood as a combination of multiple underlying topics, each characterised by a distribution of multiple words \cite{blei2003latent}.

Understanding the core concepts of LDA is essential for gaining insight into topic modelling. LDA is built upon two fundamental equations, denoted as Equation \ref{lda_0} and Equation \ref{lda_1}. For the sake of clarity, we introduce a new variable, denoted as $Q$, which signifies the relationship between each word in each document and the associated topic. $P(W|\alpha,\beta)$ represents the likelihood probability \cite{blei2003latent} of the entire document collection. The parameter $\alpha$ plays a pivotal role in controlling the mixture of topics within each document, while $\beta$ governs the distribution of words within each topic, determining which words are associated with specific topics. $z_{m,n}$ denotes the topic of the n-th word in the m-th document, $w_{m,n}$ signifies the n-th word in the m-th document, and $\beta_{z_{m,n}}$ represents the vocabulary distribution of topic $z_{m,n}$. In this research, we will not explain the mathematical derivation of LDA in depth. With core Equation \ref{lda_0} and Equation \ref{lda_1} of LDA, we aim to provide readers with an intuitive understanding of LDA's operational principles and its robust foundation as a powerful probabilistic model \cite{blei2003latent}.

\begin{equation}\label{lda_0}
P(W|\alpha,\beta) = \prod_{m=1}^{M} \int P(\theta_{m}|\alpha) Q \, d\theta_{m},
\end{equation}

\begin{equation}\label{lda_1}
Q = \prod_{n=1}^{N} \sum_{z_{m,n}} P(z_{m,n}|\theta_{m})P(w_{m,n}|\beta_{z_{m,n}}).
\end{equation}

Mahanty et al. \cite{mahanty2019} conducted an in-depth investigation into the evolution of the CE concept within the realm of academic articles. Their approach involved the utilisation of a straightforward and intuitive LDA model to compute probability distributions after topic generation. This LDA model was leveraged to analyse and compare the topics within each year. These probability distributions were then meticulously examined to discern prevalent trends in topics across the years. It is crucial to underscore that their analysis of LDA revealed an inherent limitation in the model's capacity to accurately capture the temporal patterns embedded within a corpus of academic texts. Recognising this constraint, they thoroughly explored alternative models tailored to elucidate the dynamics of temporal changes within their academic text dataset.

Among the models designed to address the temporal dimension, the Topics Over Time (TOT) model \cite{wang2006} stands out for its distinctive feature of employing beta-distributed topics, deviating from the convention of Gaussian-distributed topics typically found in Dynamic Topic Models (DTM) \cite{blei2006}. It is noteworthy that the applicability of the TOT model is limited by its constraint of permitting only a point, allowing for one instance of rising and falling trends. This constraint imposes a significant limitation on the model's ability to capture intricate and multifaceted changes in topics, such as scenarios involving a sequence of topic frequency decline, upswing, and subsequent decline \cite{chen2015}. Our text data exhibits analogous characteristics. Given that our data primarily revolve around the collection of public attention and the meticulous analysis of textual content, we anticipate the emergence of fluctuations in topic distributions over time. Hence, we opt for an initial approach utilising the fundamental LDA model to conduct a probabilistic-based topic modelling of the CE. In the research, we employ the fundamental Bag-of-Words (BoW) model \cite{zhang2010} as a foundational element in our LDA model. The BoW model focuses exclusively on the presence or absence of individual words within documents, without regard to the sequential order in which they appear. This approach treats phrases such as `Britain loves the circular economy' and `The circular economy loves Britain' as equivalent, as the model disregards the specific word order. By discovering the recurrent topics found in the corpus and assessing how common they are in each document, allows us to establish a foundational understanding of CE topics within the corpus.

Subsequently, we extend our analysis by employing other models: \nameref{subsec:corex} and \nameref{subsec:bertopic}, offering distinct methodologies for topic modelling. This multi-pronged strategy facilitates a comprehensive exploration of the dynamics surrounding public attention and its evolving facets regarding the concept of CE, enabling a deeper comprehension of attention changes within public attention.

\subsection{Correlation Explanation (CorEx) Topic Model}\label{subsec:corex}

Correlation Explanation (CorEx) is a semi-supervised \cite{van2020survey} topic model aimed at uncovering latent structures and topics within data. In contrast to traditional LDA, CorEx leverages the principle of maximum correlation from information theory, seeking to identify a set of latent topics that maximise the mutual information between the data and the topics \cite{gallagher2018corex}.

CorEx's objective is to discover the hidden topics by maximising the mutual information $I(X;T)$ between the data variables $X$ and the latent topic variables $T$. Formally, it seeks to maximise:

\begin{equation}
\max_{p(t|x)} I(X;T) = \max_{p(t|x)} \sum_{x,t} p(x,t)\log\frac{p(x,t)}{p(x)p(t)},
\end{equation}
where $p(x,t)$ is the joint distribution \cite{sklar1973random} of $X$ and $T$, and $p(x)$ and $p(t)$ are the marginal distributions \cite{stuart1955test} of $X$ and $T$, respectively. To compute $I(X;T)$, CorEx employs a KL-divergence-based approximation \cite{goldberger2003efficient}, defining a tractable lower bound:

\begin{equation}
I(X;T) \geq \mathbb{E}_{p(t|x)}\left[\sum_{x,t}p(t|x)p(x)\log\frac{p(t|x)}{r(t)}\right],
\end{equation}
where $r(t)$ is an estimate of the marginal distribution of $T$. By maximising this lower bound, CorEx can learn $p(t|x)$, the distribution of topics $T$. CorEx's strength lies in its semi-supervised nature, enabling the incorporation of prior knowledge to guide the topic discovery process, thereby yielding more interpretable and meaningful topics. In this research, we designated the words `reduce', `reuse', and `recycle' from the 3Rs policy \cite{sakai2011international} as anchor words (prior knowledge) for the CorEx model, leveraging this prior knowledge to steer the model towards discovering topics pertinent to the circular economy. By employing these anchor words as known topic words, CorEx endeavours to uncover other highly correlated topics and better capture the semantics of the circular economy.

\subsection{BERTopic}\label{subsec:bertopic}

BERTopic utilises the pre-trained language model BERT \cite{devlin2019bert} to identify topics in text data. In the research, we employ BERTopic to gain insights into public attention regarding the circular economy. The BERTopic with the class-based c-TF-IDF \cite{grootendorst2022} procedure can leverage contextual understanding of words and phrases, and identify topics that may not be apparent through traditional methods \cite{grootendorst2022}. Alamsyah et al. \cite{alamsyah2023} sheds light on the remarkable capabilities of BERTopic in the analysis of attention within textual data. In their comparative analysis, BERTopic stands out as an effective model, achieving a notable coherence score \cite{alamsyah2023} of $0.67$ when applied to user comments. This performance is particularly impressive when considering that the text inputs are often concise, aligning with the short text data paradigm, as previously discussed in the \nameref{subsec:corex}. However, they \cite{grootendorst2022, alamsyah2023} do not consider the performance of BERTopic on non-short texts (such as our The Guardian news data), nor does it examine whether BERTopic can maintain the same effect on different types of data. As we embark on the research, it is essential to acknowledge that the dataset will encompass text of varying lengths, including non-short text \cite{jipeng2019short}. This diversity in textual data prompts a natural curiosity regarding the adaptability and performance of BERTopic in handling these more extended textual inputs.

\begin{figure}[ht]
    \centering
    \includegraphics[scale=0.95]{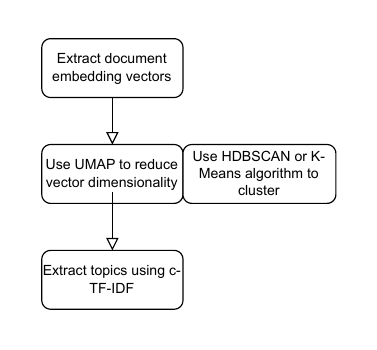}
    \caption{The process of BERTopic topic modelling}
    \label{bertopic_flowchart}
\end{figure}

The initial steps of BERTopic involve the basic operations of converting words from the English language to numeric or vector representations (Embedding Vector). As shown in Figure \ref{bertopic_flowchart}, this key process plays an important role in the BERTopic model to recognise human language. This transformation is used to encode the semantics of words into a format suitable for BERTopic model manipulation, thus facilitating our subsequent topic modelling. Uniform Manifold Approximation and Projection (UMAP) will transform the vector obtained in BERTopic into a lower dimensional representation, so as to facilitate BERTopic to understand the data more intuitively. UMAP first finds the nearest neighbours. We can specify the number of neighbours by adjusting the $n_{neighbors}$ hyper-parameter of UMAP. UMAP uses the form of constructing relations in higher dimensions. For $K$ adjacent points $n$ of a certain point $m$, there is a relationship of Equation \ref{umap_0}. Where $d_{\min(m)}$ represents the Euclidean distance \cite{dokmanic2015} from the  point to its nearest neighbour. The UMAP algorithm calculates the distance between each point $m$ and its closest neighbour, represented as $d_{\min(m)}$. Subsequently, it employs a Binary Search algorithm as described by Williams \cite{williams1976AMT} to determine the optimal value of $\sigma_{m}$ required for solving the conditional probability $p_{m|n}$. This iterative process continues until the joint probability \cite{carlton2017} $p_{mn}$ which is defined in Equation \ref{umap_1}, converges to the desired state. 

\begin{equation}\label{umap_0}
p_{m|n} = e^{-\sigma_{m} \cdot d(x_m, x_n) - d_{\min(m)}},
\end{equation}

\begin{equation}\label{umap_1}
p_{mn} = p_{m|n} + p_{n|m} - p_{m|n} \cdot p_{n|m},
\end{equation}

\begin{equation}\label{umap_2}
q_{mn} = \frac{1}{1 + a(y_m - y_n)^{2b}}.
\end{equation}

UMAP uses the representation of joint probability to represent high-latitude data. McInnes et al. \cite{mcinnes2020umap} designed a specific family of curves $\frac{1}{1+a(y^{2b})}$ to model the probability of distance in low dimensions, and Equation \ref{umap_2} directly represents the relationship between points $m$ and $n$ in the low dimensions $q_{mn}$. Both $a$ and $b$ in Equation \ref{umap_2} are hyperparameters in UMAP. In this step, UMAP solves for the hyperparameters $a$ and $b$ using $min_dist$ a representation of the minimum distance between points in low dimensions. Thus, we can obtain the initial low-dimensional matrix. 

In the last step of the whole UMAP algorithm, UMAP will use the Loss function \cite{wang2022} to make the representation of high dimensional points close to the representation of low dimensional points to achieve the effect of dimensional reduction of high-dimensional data. And according to McInnes et al. \cite{mcinnes2020umap}, UMAP will use cross-entropy \cite{mao2023CrossEntropyLF} as the Loss function instead of Kullback Leibler Divergence \cite{shlens14c}. This design can reduce the dimensional of the data by preserving the global distance when the data is moved from the high dimensional to the low dimensional \cite{mcinnes2020umap}. In this paper, we will keep the default setting of UMAP, and cross-entropy will be used as the loss function of UMAP to realise the dimensionality reduction step of BERTopic. After reducing the data's dimensional using UMAP, we can employ an appropriate approach (Hierarchical Density-Based Cluster Selection (HDBSCAN) \cite{HDBSCAN2019} or K-Means \cite{macqueen1967, jin2010}) to cluster the data effectively. HDBSCAN is a density clustering algorithm capable of identifying clusters with different shapes and sizes without the need to pre-specify the number of clusters. K-Means is a clustering algorithm that identifies $K$ clusters within a given dataset. It discovers $K$ distinct clusters, with the centre of each cluster calculated as the mean of the values contained within that cluster. The number of clusters, $K$, is specified by the user, and each cluster is described by its centroid, which is the central point of all the points in the cluster.

From the early experiments, we found that both HDBSCAN and K-Means aided BERTopic in better comprehending the clustering structure of the dimensionality-reduced data. These two methods proved valuable when dealing with more complex and noisy datasets. However, since ensuring the quality of each acquired data point in real-world scenarios is typically challenging and this paper will involve a comparative analysis of three models. Controlling the number of topics generated or clusters formed by different models can provide a clearer understanding of their performance on the same dataset. K-Means allows us to select the desired number of clusters and forces every point to belong to a cluster. Consequently, no outliers will be produced. Nevertheless, as mentioned in BERTopic \cite{grootendorst2022}, there are drawbacks. When we forcibly cluster every point, it implies that the cluster is likely to contain noise that may degrade the topic representation. Fortunately, we can leverage the built-in English stop word library in the model to remove unnecessary vocabulary, thereby enhancing the topic representation. All in all, the K-Means clustering algorithm will be the choice for this research.

\begin{equation}\label{tf}
\text{TF}_{i,j} = \frac{n_{i,j}}{\sum_{N} \cdot n_{i,j}},
\end{equation}

\begin{equation}\label{idf}
\text{IDF}(i) = \log\left(\frac{N}{N_{i}+1}\right),
\end{equation}

\begin{equation}\label{ctfidf}
W(t, c) = \text{TF}_{t,c} \cdot \log (1 + \frac{A}{t \cdot f_{t}}).
\end{equation}

BERTopic modifies the Term Frequency-Inverse document frequency (TF-IDF) algorithm \cite{tfidf2004}. As the last step of the model, it uses class-based TF-IDF (c-TF-IDF) \cite{grootendorst2022} to extract the topic words. TF-IDF is a statistical technique, the same as TF $\cdot$ IDF, wherein TF stands for term frequency, and IDF stands for inverse document frequency. In Equation \ref{tf}, $i$ represents a specific term, $j$ corresponds to a particular document, and $N$ denotes the total number of words within the document. Equation \ref{tf} allows us to compute the term frequency (TF) of a term $i$, which can also be interpreted as the probability of the term $i$ occurring in the content $j$. The IDF for a term $i$ is computed by dividing $N$ by $N_{i}$ which represents the number of documents containing the term $i$ and taking the algorithm of the resulting quotient (Equation \ref{idf}). However, in cases where the term $i$ does not appear in any document, it could lead to a division by zero issue. To prevent this, it is a common practice to $+1$ to the numerator to ensure the formula remains solvable.

From Robertson's research \cite{tfidf2004}, we understand that TF-IDF is employed to assess the significance of words. When the term frequency (TF) of the word $i$ is high in one text but low in other articles, it is deemed that this word $i$ possesses strong category-distinguishing characteristics and is suitable for classification. However, the basic structure of IDF (Inverse Document Frequency) does not take into account the semantic nuances of words or effectively capture word distribution, rendering the IDF's methodology somewhat rudimentary \cite{sun2022TextCA}. Consequently, it struggles to address issues related to word polysemy and homonymy, and as a result, the accuracy of the TF-IDF algorithm is not particularly high, especially when working with pre-classified text datasets \cite{sun2022TextCA}. Fortunately, researchers have made advancements in text representations for classification. BERTopic leverages an adapted TF-IDF model, enabling the extraction of keywords based on classified text, which is facilitated through the c-TF-IDF approach \cite{grootendorst2022, sun2022TextCA}. In Equation \ref{ctfidf}, within the context of the c-TF-IDF algorithm. $W(t, c)$ denotes the weighted score (importance degree) \cite{grootendorst2022} of word $t$ in the category $c$. $f_{t}$ represents the frequency of word $t$ across all categories, and $A$ signifies the average word count in each category. A notable departure from the traditional TF-IDF method can be observed in the IDF formula (Equation \ref{idf}).

Following the modification of the c-TF-IDF algorithm, it becomes apparent that as $f_{t}$ increases, the likelihood of the word $t$ being a common term rather than one specific to a particular category also rises. In our research, we aim to achieve results similar to those in Sun et al. \cite{sun2022TextCA}. BERTopic effectively addresses a scenario where the c-TF-IDF algorithm may inadvertently overlook keywords in similar texts, thereby enhancing the accuracy of topic modelling.

\subsection{Evaluation Metrics}\label{subsec:metrics}

\paragraph{\textbf{Coherence}}

Several metrics have been suggested for assessing the effectiveness of topic modelling techniques. For example, researchers have assessed the performance of topic models by gauging their accuracy in information retrieval, as demonstrated in prior work \cite{wei2006lda}. Additionally, statistical approaches have been employed to evaluate a topic model's predictive capability on unseen documents, typically quantified through perplexity calculations \cite{wallach2009evaluation}. However, topic coherence metrics have emerged as a predominant and widely used metric for evaluating topic models in recent years \cite{aletras2013evaluating}: For a given topic, we select the $N$ most relevant terms as representative words. We pair these $N$ representative words to form a set of bigrams $\frac{(N(N-1))}{2}$. For each bigram, denoted as $(\omega_1, \omega_2)$. For each bigram, we compute a score using the Normalised Pointwise Mutual Information (NPMI) \cite{church1990, bouma2009NormalizedM}. NPMI is a normalisation operation based on Pointwise Mutual Information (PMI) \cite{bouma2009NormalizedM, role2011pmi}. PMI between two words $\omega_1$ and $\omega_2$ is defined as:

\begin{equation}
\text{PMI}(\omega_1, \omega_2) = \log\frac{P(\omega_1, \omega_2)}{P(\omega_1)P(\omega_2)},
\end{equation}
where $P(\omega_1, \omega_2)$ is the probability of the two words co-occurring, and $P(\omega_1)$ and $P(\omega_2)$ are the individual probabilities of each word occurring. PMI measures the relationship between two words by comparing the observed co-occurrence probability with the expected probability if the words were independent. However, PMI is biased towards infrequent co-occurrences, which can lead to unintuitive results for topic coherence evaluation. To address this issue, the Normalised PMI (NPMI) is employed \cite{bouma2009NormalizedM}:

\begin{equation}
\text{NPMI}(\omega_1, \omega_2) = \frac{\text{PMI}(\omega_1, \omega_2)}{-\log P(\omega_1, \omega_2)},
\end{equation}
NPMI normalises PMI to range between -1 and 1, where 1 indicates perfect coherence, 0 indicates independence, and -1 indicates perfect incoherence. We then take the average of the scores of all bigrams within a topic as the coherence score for that topic. Finally, we average the coherence scores across all topics to obtain the overall coherence score for each model.

To further enhance the measurement of topic coherence, Röder et al. \cite{roder2015exploring} introduced the $C_{\text{NPMI}}$ metric, which extends NPMI by averaging its overall word pairs in a topic's top $N$ words. The $C_{\text{NPMI}}$ coherence score for a single topic is calculated as:

\begin{equation}
C_{\text{NPMI}} = \frac{1}{\binom{N}{2}} \sum_{i < j} \text{NPMI}(\omega_i, \omega_j),
\end{equation}
where $\omega_i$ and $\omega_j$ are the representative words of the topic. This metric effectively captures the overall semantic coherence by considering the pairwise associations between all top terms within a topic. The final coherence score for the model is then obtained by averaging the $C_{\text{NPMI}}$ scores across all topics.

By employing $C_{\text{NPMI}}$, we obtain a more robust evaluation of topic quality, as it mitigates the biases inherent in PMI and NPMI when applied to individual word pairs. This comprehensive approach to measuring coherence has been widely adopted in recent topic modelling research due to its effectiveness in reflecting human judgments of topic interpretability.

\paragraph{\textbf{Diversity}}

Topic diversity is a crucial metric for evaluating the distinctiveness and breadth of topics produced by a topic model. It assesses the extent to which topics are unique and non-overlapping, ensuring that the model captures a wide range of topics present in the corpus \cite{dieng2019dynamic}. To quantify diversity, we utilise the Inverted Rank-Biased Overlap (Inverted RBO) metric \cite{corsi2024treatment}, which measures the dissimilarity between ranked lists of words representing different topics.

Rank-Biased overlap (RBO) is a similarity measure for ranked lists that accounts for top-weightedness, meaning higher-ranked items contribute more to the overall score \cite{corsi2024treatment}. Given two topics $t_i$ and $t_j$, each represented by an ordered list of their top $N$ words, the RBO is defined as:

\begin{equation}
\text{RBO}(t_i, t_j) = (1 - p) \sum_{k=1}^{N} p^{k-1} \cdot \frac{|W_{t_i}^{(k)} \cap W_{t_j}^{(k)}|}{k},
\end{equation}
where $W_{t}^{(k)}$ is the set of the top $k$ words in topic $t$. $p$ is the persistence parameter $(0 < p < 1)$, controlling the weight decay; a higher $p$ gives more weight to lower-ranked items. To measure diversity, we compute the Inverted RBO between all pairs of topics, which reflects the dissimilarity:

\begin{equation}
\text{Inverted\ RBO}(t_i, t_j) = 1 - \text{RBO}(t_i, t_j).
\end{equation}

The overall diversity $D$ of the topic model is then calculated as the average Inverted RBO across all unique topic pairs:

\begin{equation}
D = \frac{2}{|T|(|T| - 1)} \sum_{i=1}^{|T|} \sum_{j=i+1}^{|T|} \left(1 - \text{RBO}(t_i, t_j)\right),
\end{equation}
where $|T|$ is the total number of topics. A higher $D$ value indicates greater diversity among topics, signifying that the model captures a wider array of distinct topics with minimal redundancy.

\paragraph{\textbf{Perplexity}}

This is a standard metric used to evaluate the predictive performance of probabilistic models in natural language processing \cite{blei2003latent}. It gauges how well a model predicts a sample of data, with lower perplexity values indicating better generalisation to unseen data.

In the context of topic modelling, perplexity assesses the model's ability to assign high probabilities to words in unseen documents. Formally, the perplexity $\mathcal{P}$ of a held-out test set is defined as:

\begin{equation}
\mathcal{P} = \exp\left( - \frac{\sum_{d=1}^{D} \sum_{w=1}^{N_d} \log P(w_{d,w} \, | \, \theta_d)}{\sum_{d=1}^{D} N_d} \right),
\end{equation}
where $D$ is the number of documents in the test set. $N_d$ is the number of words in document $d$. $w_{d,w}$ is the $w$-th word in document $d$. $P(w_{d,w} \, | \, \theta_d)$ is the probability of word $w_{d,w}$ given the document's topic distribution $\theta_d$. 

To compute perplexity, we ensure numerical stability by avoiding zero probabilities, typically by applying a small constant $\epsilon$ (e.g., $1 \times 10^{-10}$) to clip the probabilities:

\begin{equation}
P(w_{d,w} \, | \, \theta_d) = \max\left( P(w_{d,w} \, | \, \theta_d), \epsilon \right).
\end{equation}

This adjustment prevents the logarithm from becoming undefined due to zero probabilities. The perplexity metric effectively evaluates the exponential of the negative average log-likelihood per word, providing insight into the model's predictive uncertainty \cite{grootendorst2022}. A lower perplexity score implies that the model is more adept at predicting word occurrences in unseen documents \cite{dieng2019dynamic}.

\paragraph{\textbf{Pareto Front}}

The Pareto front represents the set of optimal solutions in Multi-objective Optimisation where no objective can be improved without degrading another \cite{van1998evolutionary, giagkiozis2014pareto}. In topic modelling, this concept helps us balance conflicting metrics such as coherence, diversity, and perplexity \cite{dieng2019dynamic}. A solution $\mathbf{x}^*$ is Pareto optimal if there is no other solution $\mathbf{x}$ such that:

\begin{equation}
\forall m, \quad f_m(\mathbf{x}) \leq f_m(\mathbf{x}^*) \quad \text{and} \quad \exists m, \quad f_m(\mathbf{x}) < f_m(\mathbf{x}^*),
\end{equation}
where $f_m$ are the objective functions representing different evaluation metrics. By mapping these solutions, we identify hyperparameter settings that offer the best trade-offs, aiding in the selection of models that align with specific performance goals.

\paragraph{\textbf{Tree-structured Parzen Estimator Application in Multi-objective Optimisation}}

The Tree-structured Parzen Estimator (TPE) is a Bayesian optimisation algorithm effective for hyperparameter tuning \cite{ozaki2022multiobjective}. While originally designed for single-objective optimisation, TPE can be adapted for multi-objective problems by optimising a scalarised objective function \cite{ozaki2020multiobjective}. In this research, we employ TPE to navigate the hyperparameter space efficiently, aiming to find models that balance coherence, diversity, and perplexity.

\begin{figure}
    \centering
    \includegraphics[width=0.75\linewidth]{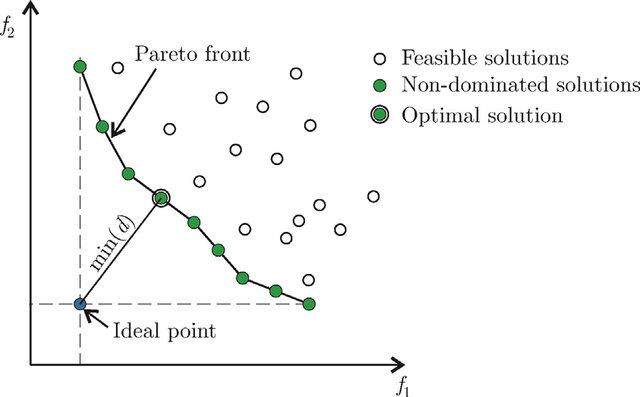}
    \caption{Schematic representation of the Pareto front, the ideal point, and the final optimal solution minimising two contradictory objectives $f_1$ and $f_2$ \cite{bre2017computational}.}
    \label{fig:ideal}
\end{figure}

Although the Pareto front provides a set of non-dominated solutions, selecting a single optimal point requires further criteria. We utilise the Ideal Point Method \cite{bre2017computational}, which defines an ideal point $\mathbf{z}^*$ composed of the optimal values for each objective:

\begin{equation}
\mathbf{z}^* = \left( \min_{\mathbf{x}} f_1(\mathbf{x}), \min_{\mathbf{x}} f_2(\mathbf{x}), \ldots, \min_{\mathbf{x}} f_M(\mathbf{x}) \right),
\end{equation}
where $f_m(\mathbf{x})$ are the objective functions. For each solution $mathbf{x}$ on the Pareto front, we calculate the Euclidean distance (schema shown in \Cref{fig:ideal}) to the ideal point:

\begin{equation}
d(\mathbf{x}) = \sqrt{ \sum_{m=1}^{M} \left( f_m(\mathbf{x}) - z^*_m \right)^2 }.
\end{equation}

We select the solution with the minimal distance $d(\mathbf{x})$ as the optimal hyperparameter configuration. This approach allows us to convert the multi-objective problem into a single-objective one, making it compatible with TPE optimisation \cite{ozaki2022multiobjective, ozaki2020multiobjective}. By integrating TPE with the Ideal Point Method, we effectively handle multiple objectives and identify hyperparameters that offer the best trade-offs among coherence, diversity, and perplexity.

\subsection{Circular Economy}\label{subsec:circular}
In recent years, the concept of the CE has become a major paradigm shift, emerging as a viable alternative to the conventional linear economic model. By focusing on the regenerative and ethical use of resources throughout their lifespan, CE aims to minimise waste and maximise resource efficiency. At its core are the principles of reducing, reusing, and recycling, synergistically creating  
an economic framework that is more ecologically friendly and sustainable \cite{korhonen2018}. 

One strand of research examined the public recognition and understanding of the concept of the CE from a broader social perspective \cite{Smol2018, Guo2017}. Some authors focused on exploring consumers’ attitudes and purchase intentions towards the CE, including preferences for remanufactured products and circularly packaged products \cite{Onete2018, Wang2018}. Furthermore, the literature analysed companies’ perspectives on the transition to a circular economy, how they incorporate circular economy initiatives into their production operations, and the obstacles they face in this process \cite{Masi2018, Liakos2019, Garcia-Quevedo2020}. To ensure broader coverage of CE-related news data, 53 keywords (Appendix \Cref{appendix:keywords}) were included in the experiment to ensure  Guardian news data can be obtained more widely.

Digitalisation has significantly reshaped a firm's production models in recent years \cite{Chauhan2022}, extending its influence beyond information technology to various aspects, including circular practices. For instance, digital technologies play a crucial role in achieving carbon neutrality by enhancing product value and reducing pollution \cite{Khan2021}. The application of digital technology not only improves the efficiency of the CE but also allows companies to broaden the scope of its applications and effectively utilise scarce resources \cite{Berg2020}. Furthermore,  integrating the circular economy concept into product and service design substantially enhances consumption potential by decoupling growth from resource availability \cite{Wang2018}. Previous research has identified a positive relationship between digitalisation and the development of the CE \cite{Berg2020}. Tools such as big data analytics and the Internet of Things (IoT), when combined with digital technology, enable monitoring and tracking of product life cycle data, thus minimising waste and enhancing recycling rates \cite{Esmaeilian2020}. Artificial intelligence enhances productivity and promotes CE development through optimised processes and real-time data analysis \cite{Berg2020}. Additionally, machine learning anticipates uncertainties in the design and production process, identifying deficiencies in the CE system through real-time monitoring \cite{Sundui2021}. Therefore, analysing the circular economy from a digital perspective can assist companies in achieving a low-carbon transformation and effectively capturing public views and attitudes towards the circular economy.

%% file: section3_clean.tex
\section{Methodology}\label{sec:methodology}

\subsection{Framework}

\begin{figure}
    \centering
    \includegraphics[scale=0.40]{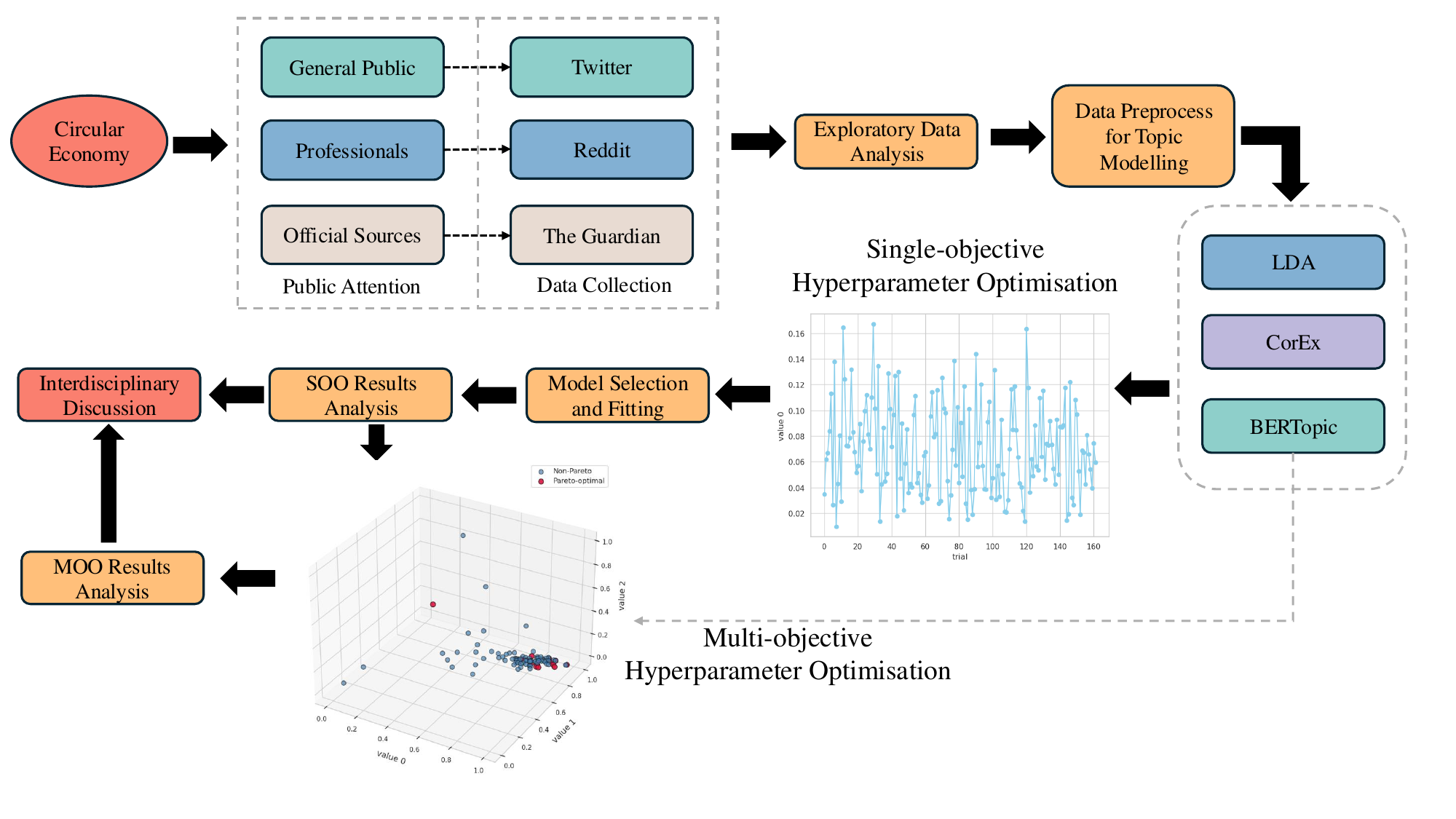}
    \caption{The proposed framework employs single-objective and multi-objective hyperparameter optimisation on topic modelling in the context of the circular economy. SOO and MOO respectively denote single-objective and multi-objective optimisation.}
    \label{fig:framework}
\end{figure}

\Cref{fig:framework} depicts the proposed framework for this interdisciplinary research, which employs topic modelling with twin hyperparameter optimisations and places particular emphasis on investigating public attention regarding the CE.

Firstly, we identified the research context, CE. To explore public attention regarding CE, we divided the sources into three groups: the general public, professionals, and official entities. Then, we collected textual posts from Twitter, Reddit, and The Guardian platforms to reflect the attention of these three distinct strata of the public. It is worth noting that, to ensure compliance with data regulations, we utilised official APIs to retrieve the relevant data. More details regarding the datasets can be found in \Cref{sec:datasets}. 

Next, we data.Next,an  we conducted exploratory data analysis to gain a preliminary understandisetdata landscape before performing necessary datad As shown in \Cref{fig:framework}, LDA, CorEx, and BERTopic represent the three distinct topic models used in this research, each chosen for its specific motivation and approach to topic modelling. These models were subjected to single-objective hyperparameter optimisation using grid search, allowing for a relatively thorough, simple, and systematic exploration of the hyperparameter space to identify the optimal values. Coherence was selected as the objective metric, as it is a well-known and commonly used measure of topic quality in topic modelling.

Many advanced approaches have been proposed to automate the hyperparameter optimisation process, such as CMA-ES \cite{hansen2016cma} and TPE \cite{bergstra2011}. However, these approaches usually perform well and exhibit their strengths on relatively large hyperparameter search spaces. According to our designed hyperparameter search space, we consider grid search a promising option as it can exhaustively explore all solutions explicitly.

However, single-objective hyperparameter optimisation can potentially overlook other critical aspects, such as the diversity and perplexity of the generated topics. Therefore, we proposed a twin hyperparameter optimisation framework. We selected the best-performing topic modelling approach from the single-objective optimisation results to further apply multi-objective optimisation using the Tree-structured Parzen Estimator algorithm on a finer hyperparameter grid. We then analysed the two sets of topics generated from the models optimised by both single-objective optimisation (SOO) and multi-objective optimisation (MOO), with the latter serving as auxiliary information, offering an alternative perspective to support our investigation of CE. Additional details on SOO and MOO can be found in \Cref{section:hpofortopicmodelling}.

\subsection{Datasets}\label{sec:datasets}
The CE is a broad topic that involves various stakeholders, including the general public, professionals, and officials. Analysing the diverse opinions and interests in CE across these groups is essential for understanding public attention, which necessitates collecting data from a wide spectrum of society. To accomplish this, textual posts were collected from Twitter and Reddit, representing the perspectives of the general public and professionals, respectively. Additionally, news articles from The Guardian were gathered to analyse official opinions and attention regarding CE. We collected a total of 22,106 documents related to the CE and related keywords from three different sources, strictly following each platform's API usage policy, and each dataset was based on a fixed time range of 1999 to 2023.

The process of data scraping for this research can be summarised in the following steps:

\begin{enumerate} 

\item Search for suitable websites containing data related to CE from various segments of the public to form a comprehensive dataset, ensuring that subsequent analysis can cover a wide range of perspectives on CE.
\item Check the policies of the selected websites to ensure that data scraping is permitted for research purposes and that the process complies with their regulations. 
\item Follow the university's rules to submit an ethics approval application. 
\item Investigate the usage of the APIs offered by the websites. Data scraping can be challenging, especially for websites that store large amounts of data, as searching for user-specified data can be time-consuming. Additionally, different websites may design different APIs with varying parameters (e.g., keywords, time spans) to support efficient data scraping. Therefore, it is important to be familiar with the API usage to ensure that the downloaded data meets expectations. \end{enumerate}

In machine learning, data preprocessing is a critical step for models to achieve satisfactory performance. This step is also vital for topic modelling because the quality of the original scraped data is often poor and not suitable for direct use in topic models. For example, website links commonly appear in social media posts and news articles; however, they are less informative for topic modelling and may adversely affect model performance. In this research, data preprocessing primarily involves several tasks: stopword removal, punctuation removal, email and URL removal, special symbol removal, lemmatisation, and checking whether the text length exceeds the model's limit.

Data were collected in JSON format from three primary sources: The Guardian, Reddit, and Twitter. From The Guardian, we extracted full news articles including titles, sections, publication dates, URLs, and complete content. Reddit data comprised post titles, creation timestamps, URLs, and main content (self-text), excluding comments. Twitter data included both original tweets and retweets, encompassing user screen names, creation timestamps, tweet IDs, and full tweet text. Comments and replies on Reddit and Twitter were not included in our dataset to maintain focus on primary content related to circular economy topics.


\subsection{Hyperparameter Optimisation for Topic Modelling}\label{section:hpofortopicmodelling}

Hyperparameters in machine learning are parameters that must be set before training or fitting a model. In other words, a machine learning algorithm is configured with a specific set of hyperparameter values at the time of instantiation. These values control the behavior of the algorithm and influence key aspects such as model complexity and convergence. Consequently, different sets of hyperparameter values can lead to varying results. Hyperparameter selection is inherently an iterative process of trial and error, and it typically relies on the evaluation of a quantitative metric.

\subsubsection{SOO}
Single-objective hyperparameter optimisation is a common step in machine learning research, as most tasks in machine learning follow standard evaluation criteria. For example, in classification tasks, accuracy is often prioritised. In our research, to compare different sets of hyperparameter values, we used the coherence ($C_{\text{NPMI}}$) value, which ranges from -1 to 1; a value of 1 indicates the best coherence.

The SOO of topic modeling hyperparameters is defined in \Cref{eq:soo}.
\begin{equation}
\label{eq:soo}
    \boldsymbol{\lambda^*}=\underset{\boldsymbol{\lambda} \in \boldsymbol{\Lambda}}{\operatorname{argmin}} \mathbf{V}\left(\mathcal{L}, \boldsymbol{\mathcal{A}}_{\boldsymbol{\lambda}}, D\right),
\end{equation}
given a textual dataset $D$, the goal is to find the optimal hyperparameter solution $\boldsymbol{\lambda^*}$, where $\mathbf{V}\left(\mathcal{L}, \boldsymbol{\mathcal{A}}_{\boldsymbol{\lambda}}, D\right)$ measures the quality of the topics generated by the topic modeling approach $\boldsymbol{\mathcal{A}}$ with a hyperparameter solution $\boldsymbol{\lambda}$ on dataset $D$. The quality is determined by a predefined objective function $\mathcal{L}$. $\boldsymbol{\lambda}$ is a vector of length $N$, representing a hyperparameter solution with $N$ hyperparameters (e.g., the number of top words, ngram range), and $\boldsymbol{\Lambda} = \Lambda_1 \times \Lambda_2 \times \ldots \Lambda_N$ is the search space. 

\begin{figure}
    \centering
    \includegraphics[width=0.3\linewidth]{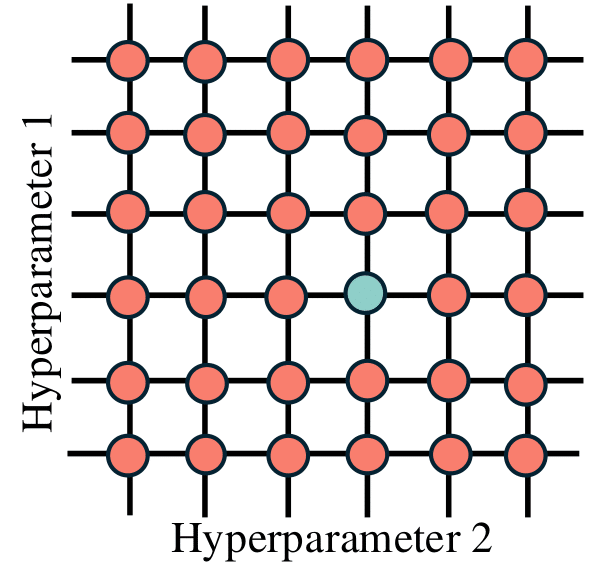}
    \caption{Grid Search}
    \label{fig:gs}
\end{figure}

To search for the optimal solution in the single-objective case, we propose using grid search in this research. An example of grid search for two hyperparameters is shown in \Cref{fig:gs}. For each combination (represented as a point in the example) of hyperparameters, the model is trained and evaluated using a chosen performance metric. This approach facilitates the identification of the optimal hyperparameter set by comparing the performance across all combinations. The primary advantage of grid search is its simplicity and thoroughness. It ensures that all possible combinations of the specified hyperparameter values are explored, guaranteeing that the optimal set within the grid is found. This exhaustive approach is particularly beneficial when the hyperparameter space is relatively small, as it allows for a comprehensive evaluation of model performance across all parameter settings. In our research, we limited the size of the grid search space to expedite the discovery of the optimal solution, but the search space was primarily designed based on literature review to ensure effectiveness.

\subsubsection{MOO}
As previously mentioned, SOO has inherent limitations. Unlike other machine learning tasks, evaluating generated topics in topic modelling involves various metrics and approaches, making it a more complex and subjective process. To address this issue, we propose applying multi-objective hyperparameter optimisation (MOO) to the hyperparameters of topic models to provide additional supportive information through another set of topics with different constraints.

MOO is a process used to optimise two or more conflicting objectives simultaneously. Unlike single-objective optimisation, where a single criterion is maximised or minimised, MOO seeks to find a balance among multiple objectives, often resulting in a set of trade-off solutions known as Pareto optimal solutions. Each solution in the Pareto front represents a situation where improving one objective would lead to a deterioration in another.

The MOO of topic modelling hyperparameters is defined in \Cref{eq:moo}. 
\begin{equation}
\label{eq:moo}
  P =\{\boldsymbol{\lambda}_1^*, \boldsymbol{\lambda}_2^*, ... \}=\underset{\boldsymbol{\lambda} \in \boldsymbol{\Lambda}}{\operatorname{argmin}} \mathbf{V}\left(\mathcal{L}_1, \mathcal{L}_2, \mathcal{L}_3, \mathcal{A}_{\boldsymbol{\lambda}}, D\right),
\end{equation}
given a textual dataset $D$, the goal is to find the Pareto set $P$ of optimal solution $\{\boldsymbol{\lambda}_1^*, \boldsymbol{\lambda}_2^*, ... \}$, where $ \mathbf{V}\left(\mathcal{L}_1, \mathcal{L}_2, \mathcal{L}_3, \mathcal{A}_{\boldsymbol{\lambda}}, D\right)$ measures the quality of the topics generated by the topic modelling approach $\boldsymbol{\mathcal{A}}$ with a hyperparameter solution $\boldsymbol{\lambda}$ on dataset $D$. The quality is determined by predefined objective functions $\mathcal{L}_1, \mathcal{L}_2$ and $\mathcal{L}_3$ which respectively correspond to coherence, diversity, and perplexity in our research. $\boldsymbol{\lambda}$ is a vector of length $N$, representing a hyperparameter solution with $N$ hyperparameters (e.g., the number of top words, ngram range), and $\boldsymbol{\Lambda} = \Lambda_1 \times \Lambda_2 \times \ldots \Lambda_N$ is the search space. 

To search the Pareto set, we proposed to use a Tree-structured Parzen Estimator (TPE). TPE \cite{bergstra2011} is a probabilistic model-based optimisation algorithm often used for hyperparameter tuning. TPE models the search space as two distributions: one for promising hyperparameter values (those expected to perform well) and another for less promising values \cite{yuan2021systematic}. It iteratively refines the search by focusing on the promising regions, making TPE more efficient than grid or random search, particularly when dealing with complex search spaces. MOO-TPE \cite{ozaki2022multiobjective} is an extension of the TPE algorithm that efficiently handles multiple conflicting objectives. MOO-TPE leverages the concept of Pareto dominance to find  better nondomination ranked solutions to form the promising set. However, in MOO, the results is a set of solutions which does not fit our research, as our target is to find one hyperparameter solution. Therefore, in our research, we proposed to use an approach based ideal point \cite{bre2017computational} concept to find the best hyperparameter solution, the approach is defined as the \Cref{eq:idealpoint}.

\begin{equation}
\label{eq:idealpoint}
      \boldsymbol{\lambda}_{best} = \sum_{n = 1 }^{|P|} \mathbf{U} (\boldsymbol{\lambda_i}, \boldsymbol{s(\lambda}^{*}_{n}\boldsymbol{)}).
\end{equation}

The vector $\boldsymbol{\lambda_i}$ represents the ideal point, also known as the ideal solution. For example, in our research, the ideal point is $[1,1,1]$; each dimension corresponds to one objective (e.g., coherence, diversity, and perplexity), and each value represents the best possible value for that objective. $\mathbf{U}$ is the distance function used to quantify the distance between the solutions $\boldsymbol{\lambda}^{*}_n$ within the Pareto set $P$ and $\boldsymbol{\lambda_i}$. In this research, we used the Euclidean distance, inspired by the work presented in \cite{bre2017computational}. It is worth noting that the value ranges of different objective functions may vary, so all values need to be rescaled by the function $\boldsymbol{s}$ into the same range before calculating the distances. In our research, $\boldsymbol{s}$ is implemented using MinMaxScaler, scaling each feature to $[0,1]$.


In this research, we selected hyperparameters to be optimised for three models according to the suggestions from original papers: BERTopic \cite{devlin2019bert}, CorEx \cite{gallagher2018corex}, and LDA \cite{wei2006lda}. We have decided to collaboratively use BERTopic, CorEx, and LDA for topic modelling experiments in our research, based on their complementary analytical strengths, which together provide a robust methodological framework for dissecting complex changes in public attention. BERTopic leverages the powerful capabilities of language model architectures, enabling it to handle vast amounts of unstructured text with a high degree of semantic understanding. This allows BERTopic to extract thematic clusters that reflect intricate details of public attention on the CE. CorEx employs an information-theoretic approach, capable of identifying subtle patterns and correlations within the data \cite{gallagher2018corex}. This makes the CorEx model particularly valuable in parsing the thematic structures behind public discourse, thus complementing the depth of BERTopic through meticulous detection of latent topics. LDA offers a probabilistic perspective on topic modelling, providing a broad overview of the topic distributions within the data. LDA's strength lies in its ability to generalise across large texts, thus enabling a macro-level understanding of the popularity and shifts concerning CE topics. These three methods ensure a comprehensive exploration of topic distributions, capture the granularity and breadth of public discourse, and offer a multifaceted understanding of public attention towards CE on different online platforms.

Furthermore, the respective search space is defined with the following considerations: (1). Try to make the search space size of the three models equal. (2). We used the model's default or suggested values as central values to create search spaces by consecutively adding or subtracting a certain number of fixed values. (3). We made a few adjustments for some special cases to refine the search space. For example, in BERTopic, `\emph{n\_cluster}' denotes the number of topics that a model needs to generate. This search space starts from 2 because we consider 2 to be the smallest number of topics. The hyperparameters and search spaces are summarised in Table \ref{tab:hpsandsearchspace}, and each model used the same search space for three datasets.

It is noted that random search and grid search are two popular approaches because they are simple and easy to implement. Table \ref{tab:hpsandsearchspace} outlines the search spaces we used, which were determined and suggested by relevant literature. These search spaces are relatively small compared to those typically encountered in deep learning research. Additionally, the three models employed in our work have lower computational complexity compared to large deep learning models. As a result, we opted for a grid search to conduct an exhaustive search and identify the optimal hyperparameters. One drawback of random search, in contrast to grid search, is the uncertainty introduced by its inherent randomness.

Moreover, it is worth drawing attention again to the fact that topic modelling is a type of unsupervised learning. No single quantitative metric can comprehensively evaluate the results, pointing out the importance of human evaluation and multi-objective optimisation. Therefore, in our research, we will begin by comparing the results in \Cref{section:soo} with human interventions. Subsequently, we will utilise the selected hyperparameter optimisation results to conduct further analysis (with visualisation) in the context of CE in Sections \ref{sec:official} $\sim$ \ref{sec:thepublic}.

\begin{table}
\centering
\caption{The hyperparameters selected to be optimised and their corresponding search spaces for SOO.}
\resizebox{0.8\textwidth}{!}{
\begin{tabular}{c|cc} 
\toprule
                          & Hyperparameter   & Search Space                                \\ 
\hline
\multirow{3}{*}{BERTopic} & n\_gram          & {[}1,3]                                     \\
                          & n\_clusters      & {[}2, 5, 10, 15, 20, 25]                    \\
                          & n\_components    & {[}5, 10, 15]                               \\
                          & n\_neighbors     & {[}10, 15, 20]                              \\ 
\hline
\multirow{2}{*}{CorEx}    & anchor\_strength & {[}1, 1.5, 2, 2.5, 3, 3.5, 4, 4.5, 5, 5.5]  \\
                          & n\_hidden        & {[}3, 5, 10, 15, 20, 25]                    \\ 
\hline
\multirow{3}{*}{LDA}      & n\_topics        & {[}2, 5, 10, 15, 20, 25]                    \\
                          & alpha            & {[}0.04, 0.05, 0.07, 0.1, 0.2, 0.5]         \\
                          & eta              & {[}0.04, 0.05, 0.07, 0.1, 0.2, 0.5]         \\
\bottomrule
\end{tabular}}
\label{tab:hpsandsearchspace}
\end{table}

%% file: section4_clean.tex
\section{Experiments and Results}\label{sec:experiments}

\subsection{Best Hyperparameter Values from SOO} \label{section:soo}
The results of SOO for three selected models (BERTopic, CorEx, and LDA) on three datasets (The Guardian, Reddit, and Twitter) are shown in \Cref{tab:besthps}. The coherence values ($C_{\text{NPMI}}$) presented in \Cref{tab:besthps} indicate that BERTopic outperformed the other two models on two of three datasets. Subsequently, we conducted a qualitative analysis of the topics produced by each model. This involved examining the top representative words for each topic to evaluate their interpretability and relevance. The goal was to observe whether the models, when optimally tuned, could generate meaningful and distinct topics that align with the semantic structures present in the data.

\begin{table}[ht]
\centering
\caption{Best Hyperparameter Values and Corresponding Coherence Scores for BERTopic, CorEx, and LDA on Three Datasets from SOO. $\uparrow$ means higher is better, $\downarrow$ means lower is better. \textbf{Bold} values represent the best performance for each dataset. The $C_{\text{NPMI}}$ metric ranges from $-1$ to $1$. A value closer to $1$ indicates higher topic coherence, meaning the words within topics are more semantically related. A value of $0$ suggests no mutual information between words, while negative values indicate incoherence.}
\resizebox{0.75\textwidth}{!}{
\begin{tabular}{c|c|c|c} 
\toprule
\multicolumn{1}{c}{\textbf{Model}} & \multicolumn{1}{c}{\textbf{Dataset}} & \multicolumn{1}{c}{\textbf{Best Parameters}}                                                                    & \textbf{Best Coherence ($C_{\text{NPMI}}$) $\uparrow$}  \\ 
\hline
\multirow{3}{*}{BERTopic}          & The Guardian                         & \begin{tabular}[c]{@{}c@{}}n\_gram=2.0, \\ n\_clusters=2, \\ n\_components=5, \\ n\_neighbors=20\end{tabular}   & \textbf{0.1669}                    \\
                                   & Reddit                               & \begin{tabular}[c]{@{}c@{}}n\_gram=2.0, \\ n\_clusters=2, \\ n\_components=15, \\ n\_neighbors=20\end{tabular}  & \textbf{-0.0609}                   \\
                                   & Twitter                              & \begin{tabular}[c]{@{}c@{}}n\_gram=2.0, \\ n\_clusters=10, \\ n\_components=10, \\ n\_neighbors=10\end{tabular} & 0.1354                             \\ 
\hline
\multirow{3}{*}{CorEx}             & The Guardian                         & \begin{tabular}[c]{@{}c@{}}anchor\_strength=5.5, \\ n\_hidden=20\end{tabular}                                   & 0.1244                             \\
                                   & Reddit                               & \begin{tabular}[c]{@{}c@{}}anchor\_strength=1.5, \\ n\_hidden=3\end{tabular}                                    & -0.1428                            \\
                                   & Twitter                              & \begin{tabular}[c]{@{}c@{}}anchor\_strength=1.0, \\ n\_hidden=5\end{tabular}                                    & \textbf{0.1538}                    \\ 
\hline
\multirow{3}{*}{LDA~}              & The Guardian                         & \begin{tabular}[c]{@{}c@{}}n\_topics=5.0, \\ alpha=0.5, \\ eta=0.5\end{tabular}                                 & -0.0451                            \\
                                   & Reddit                               & \begin{tabular}[c]{@{}c@{}}n\_topics=2.0, \\ alpha=0.5, \\ eta=0.04\end{tabular}                                & -0.0671                            \\
                                   & Twitter                              & \begin{tabular}[c]{@{}c@{}}n\_topics=2.0, \\ alpha=0.5, \\ eta=0.2\end{tabular}                                 & 0.0194                             \\
\bottomrule
\end{tabular}}
\label{tab:besthps}
\end{table}

\subsection{Best Hyperparameter Values from MOO}

As discussed in \Cref{section:soo}, the SOO focused solely on maximising the coherence metric ($C_{\text{NPMI}}$) across three models. While this approach revealed that \textbf{BERTopic} consistently outperformed other models on two datasets, the researchers increasingly recognise the limitations of relying on a single metric for topic modelling evaluation. Trade-offs often exist between multiple objectives, such as coherence, topic diversity, and perplexity. To address this, we extended our optimisation to a MOO method to explore deeply.

The decision to apply MOO only to BERTopic was driven by several factors. In the summary, BERTopic has shown better performance under SOO, making it the most promising candidate for further optimisation. Unlike LDA and CorEx, BERTopic's integration with the BERT language model allows for more complex semantic understanding, making it more adaptable to balancing multiple objectives simultaneously. From our perspective, BERTopic is designed to integrate with the language model, and is capable of handling texts from various scenarios, such as news articles and social media posts. Therefore, BERTopic provides a richer search space for MOO, allowing for fine-tuning across different dimensions.

On the other hand, CorEx and LDA did not demonstrate sufficient flexibility or performance gains in SOO to justify their inclusion in MOO as results showed. Compared to BERTopic, the other two approaches are less flexible. BERTopic can be considered a BERT-based topic modelling framework, allowing various computational components to be changed or their hyperparameters adjusted, providing greater flexibility to handle different tasks.

\begin{table}
\centering
\caption{Best Hyperparameter Values and Corresponding Scores for BERTopic on MOO (with Ideal Point-Based Process). $\uparrow$ means higher is better, $\downarrow$ means lower is better. \textbf{Bold} values represent the best performance for each dataset. Diversity ranges from $0$ to $1$. Values closer to $1$ signify greater diversity among topics, indicating that topics are more distinct from one another. Perplexity ranges from $1$ to positive infinity. Lower perplexity values indicate better model performance, with values closer to $1$ being optimal.}
\resizebox{0.9\textwidth}{!}{
\begin{tabular}{c|c|c|ccc} 
\toprule
\textbf{Model}            & \textbf{Dataset} & \textbf{Best Parameters}                                                                                      & \textbf{$C_{\text{NPMI}}$} $\uparrow$ & \textbf{Diversity} $\uparrow$ & \textbf{Perplexity} $\downarrow$ \\ 
\hline
\multirow{3}{*}{BERTopic} & The Guardian     & \begin{tabular}[c]{@{}c@{}}n\_gram=1, \\ n\_clusters=20, \\ n\_components=11, \\ n\_neighbors=19\end{tabular} & \textbf{0.1381}  & \textbf{0.9886}    & \textbf{1.4469}      \\ 
\cline{2-6}
                          & Reddit           & \begin{tabular}[c]{@{}c@{}}n\_gram=1, \\ n\_clusters=14, \\ n\_components=13, \\ n\_neighbors=15\end{tabular} & \textbf{-0.2627} & \textbf{0.8558}    & \textbf{8.1827}      \\ 
\cline{2-6}
                          & Twitter          & \begin{tabular}[c]{@{}c@{}}n\_gram=1, \\ n\_clusters=8, \\ n\_components=7, \\ n\_neighbors=15\end{tabular}   & \textbf{-0.0111} & \textbf{0.9486}    & \textbf{74.9202}     \\
\bottomrule
\end{tabular}}
\label{tab:best_pareto}
\end{table}

By conducting MOO, we were able to identify optimal hyperparameter sets that strike a balance between these competing objectives. \Cref{tab:best_pareto} presents the results from the best-performing models on The Guardian, Reddit and Twitter datasets, with optimal values for $C_{\text{NPMI}}$, diversity, and perplexity.

For the \textbf{The Guardian} dataset, the model reached a high $C_{\text{NPMI}}$ score of 0.1381, paired with a notable diversity score of 0.9886 and a low perplexity of 1.4469, indicating a well-rounded performance in formal content. In contrast, the \textbf{Reddit} dataset showed a lower $C_{\text{NPMI}}$ of -0.2627, though the model maintained strong diversity (0.8558) and handled informal text reasonably well, with a perplexity of 8.1827. On the \textbf{Twitter} dataset, BERTopic achieved a near-neutral $C_{\text{NPMI}}$ (-0.0111) with high diversity (0.9486), but a perplexity of 74.9202 reflects the challenges posed by short, context-specific tweets. Overall, BERTopic consistently performs well across multiple datasets, particularly in capturing diverse topics, even where coherence is more difficult to maintain.

\subsection{The Guardian}

\subsubsection{SOO} \label{sec:theguardian}

\begin{table}
\centering
\caption{The topic terms over 2 topics obtained from the BERTopic model with the best hyperparameter values on the Guardian dataset.}
\resizebox{\textwidth}{!}{
\begin{tabular}{c|l} 
\bottomrule
\multicolumn{1}{c}{Topic} & \multicolumn{1}{c}{Topic Term}                                                      \\ 
\hline
0                         & energy, gas, uk, fuel, carbon, emission, increase, power, plan, rise                \\
1                         & australia, australian, nsw, minister, leader, labor, continue, say, question, need  \\
\bottomrule
\end{tabular}}
\label{tab:besthpsbertopicguardian}
\end{table}

\begin{table}
\centering
\caption{The topic terms over 20 topics obtained from the CorEx model with the best hyperparameter values on the Guardian dataset.}
\resizebox{\textwidth}{!}{
\begin{tabular}{c|l} 
\bottomrule
Topic & \multicolumn{1}{c}{Topic Term}                                                                              \\ 
\hline
0     & reduce, emission, energy, carbon, gas, climate, renewable, fuel, fossil, greenhouse                         \\
1     & reuse, design, fashion, clothe, designer, brand, shop, buy, glass, kitchen                                  \\
2     & recycle, plastic, waste, recycling, landfill, product, food, material, packaging, bottle                    \\
3     & year, time, come, photograph, day, week, month, tell, high, early                                           \\
4     & minister, election, party, leader, prime, vote, parliament, political, liberal, speak                       \\
5     & water, environmental, specie, environment, forest, wildlife, natural, land, biodiversity, river             \\
6     & update, bst, gmt, australian, australia, today, morning, labor, albanese, aap                               \\
7     & police, military, russian, war, court, crime, russia, criminal, ukraine, attack                             \\
8     & covid, 19, coronavirus, virus, vaccine, pandemic, infection, lockdown, outbreak, vaccination                \\
9     & idea, century, society, kind, sense, culture, history, inspire, modern, bear                                \\
10    & man, love, old, game, fan, star, player, play, season, ball                                                 \\
11    & think, thing, like, know, look, want, ask, feel, way, good                                                  \\
12    & government, report, say, increase, policy, state, support, public, plan, new                                \\
13    & price, economy, inflation, tax, cost, economic, rise, bank, financial, growth                               \\
14    & company, industry, market, business, investment, technology, consumer, sector, sustainable, sustainability  \\
15    & trump, biden, republican, donald, president, joe, american, administration, washington, americans           \\
16    & health, child, people, school, hospital, death, care, woman, family, medical                                \\
17    & tory, labour, conservative, uk, secretary, boris, sunak, johnson, rishi, chancellor                         \\
18    & country, world, global, international, china, africa, germany, brazil, india, summit                        \\
19    & city, local, resident, town, road, mile, park, area, car, travel                                            \\
\bottomrule
\end{tabular}}
\label{tab:besthpscorexguardian}
\end{table}

\begin{table}
\centering
\caption{The topic terms over 5 topics obtained from the LDA model with the best hyperparameter values on the Guardian dataset.}
\resizebox{\textwidth}{!}{
\begin{tabular}{c|l}
\bottomrule
\multicolumn{1}{c}{Topic} & \multicolumn{1}{c}{Topic Term}                                                 \\ 
\hline
0                         & say, water, year, people, health, new, area, city, environmental, environment  \\
1                         & year, like, £, time, photograph, car, good, go, day, new                       \\
2                         & say, government, energy, climate, year, uk, new, emission, £, change           \\
3                         & say, bst, gmt, people, update, government, minister, australia, new, party~    \\
4                         & work, people, say, year, business, company, need, like, way, use               \\
\bottomrule
\end{tabular}}
\label{tab:besthpsldaguardian}
\end{table}

In \Cref{tab:besthps}, BERTopic achieved the best performance with a coherence value of 0.1669. To further investigate the performances of the three models, \Cref{tab:besthpsbertopicguardian} to \Cref{tab:besthpsldaguardian} show the topics generated by three different models on the Guardian dataset. From our perspective, LDA topics are not desirable because the word `say' appears and dominates in its three topics, but it is less meaningful for CE. Meanwhile, BERTopic and CorEx are relatively good; we can clearly see words such as `energy', `gas', and `power', which are closely related to CE. Upon analysing the topics, we observed that several Australian-related terms appeared prominently. This is due to \emph{The Guardian}'s inclusion of articles from its Australian news section via the API. Recognising that Australia has a high level of engagement with the circular economy, we retained these articles to enrich our analysis. This inclusion aligns with our objective of capturing a global perspective on public attention towards the CE.

\begin{table}
\centering
\caption{The top 5 hyperparameter sets for BERTopic on the Guardian dataset.}
\resizebox{\textwidth}{!}{
\begin{tabular}{r|r|rrrr}
\bottomrule
\multicolumn{1}{l|}{number} & \multicolumn{1}{l|}{$C_{\text{NPMI}}$} & \multicolumn{1}{l}{params\_n\_clusters} & \multicolumn{1}{l}{params\_n\_components} & \multicolumn{1}{l}{params\_n\_gram} & \multicolumn{1}{l}{params\_n\_neighbors}  \\ 
\hline
1                           & 0.16694393                 & 2                                       & 5                                         & 2                                   & 20                                        \\
2                           & 0.16458788                 & 2                                       & 5                                         & 3                                   & 10                                        \\
3                           & 0.16329703                 & 5                                       & 15                                        & 3                                   & 15                                        \\
4                           & 0.14380017                 & 20                                      & 15                                        & 1                                   & 10                                        \\
5                           & 0.13858814                 & 2                                       & 5                                         & 3                                   & 20                                        \\
\bottomrule
\end{tabular}}
\label{tab:bertopicgstop5}
\end{table}

However, there are only two topics generated by BERTopic, which is less informative for our subsequent analysis. Therefore, we revisited the BERTopic hyperparameter optimisation results, expecting to find a set of hyperparameter values that can achieve a relatively high coherence value and produce informative topics at the same time. We list the top 5 hyperparameter sets in Table \ref{tab:bertopicgstop5} obtained by hyperparameter optimisation (HPO) on BERTopic, referring to coherence values. We found that the number 4 hyperparameter set in Table \ref{tab:bertopicgstop5} has the same number of topics as CorEx, and the coherence value 0.14380017 is higher than that of CorEx (0.1244). Therefore, we selected the hyperparameters of BERTopic from the number 4 hyperparameter set for our subsequent analysis. The new topics are shown in Table \ref{tab:selectedhpsbertguadian}.

\begin{table}
\centering
\caption{The topic terms over 20 topics obtained from the BERTopic model with the alternative hyperparameter values on the Guardian dataset.}
\resizebox{\textwidth}{!}{
\begin{tabular}{c|l}
\bottomrule
Topic & \multicolumn{1}{c}{Topic Term}                                                                         \\ 
\hline
0     & australia, australian, nsw, election, labor, bst, senator, coalition, need, leader                              \\
1     & renewable, carbon, energy, emission, international, environment, world, resource, goal, climate                 \\
2     & energy, renewable, uk, gas, fuel, carbon, solar, britain, emission, electricity                                 \\
3     & mp, corbyn, eu, britain, tory, uk, brexit, labour, conservative, british                                        \\
4     & recycling, recycle, recyclable, plastic, waste, reuse, circular, packaging, environment, environmental          \\
5     & sustainable, environment, farming, green, land, grow, plan, uk, farmer, produce                                 \\
6     & sustainability, sustainable, environment, approach, resource, ethical, consumer, environmental, growth, future  \\
7     & australia, energy, renewable, nsw, australian, 2050, emission, gas, fuel, carbon                                \\
8     & candidate, biden, trump, election, voter, republicans, official, democrat, campaign, senator                    \\
9     & vaccine, lockdown, covid, vaccination, outbreak, vaccinate, quarantine, coronavirus, gmt, health                \\
10    & ftse, eurozone, gmt, trading, inflation, market, uk, price, investor, european                                  \\
11    & feel, moment, want, people, audience, picture, say, tell, die, like                                             \\
12    & airbnb, rent, sharing, share, create, offer, site, host, idea, uber                                             \\
13    & player, sport, goal, liverpool, league, kick, chelsea, want, arsenal, club                                      \\
14    & ev, tesla, car, electric, charge, nissan, vehicle, emission, petrol, driver                                     \\
15    & epa, pfas, chemical, toxic, federal, environmental, lead, pesticide, emission, official                         \\
16    & ukraine, russia, ukrainian, putin, kremlin, eu, world, russian, nato, kyiv                                      \\
17    & samsung, smartphone, tablet, android, device, price, app, screen, ipad, fold                                    \\
18    & gambling, gamble, betting, gambler, bet, casino, bookie, punter, bookmaker, gaming                              \\
19    & pga, golfer, golf, fairway, tournament, mickelson, leaderboard, spieth, championship, shoot                     \\
\bottomrule
\end{tabular}}
\label{tab:selectedhpsbertguadian}
\end{table}

\paragraph{Visualisation and Analysis} \label{sec:official}

\begin{figure}
    \centering
    \includegraphics[scale=0.35]{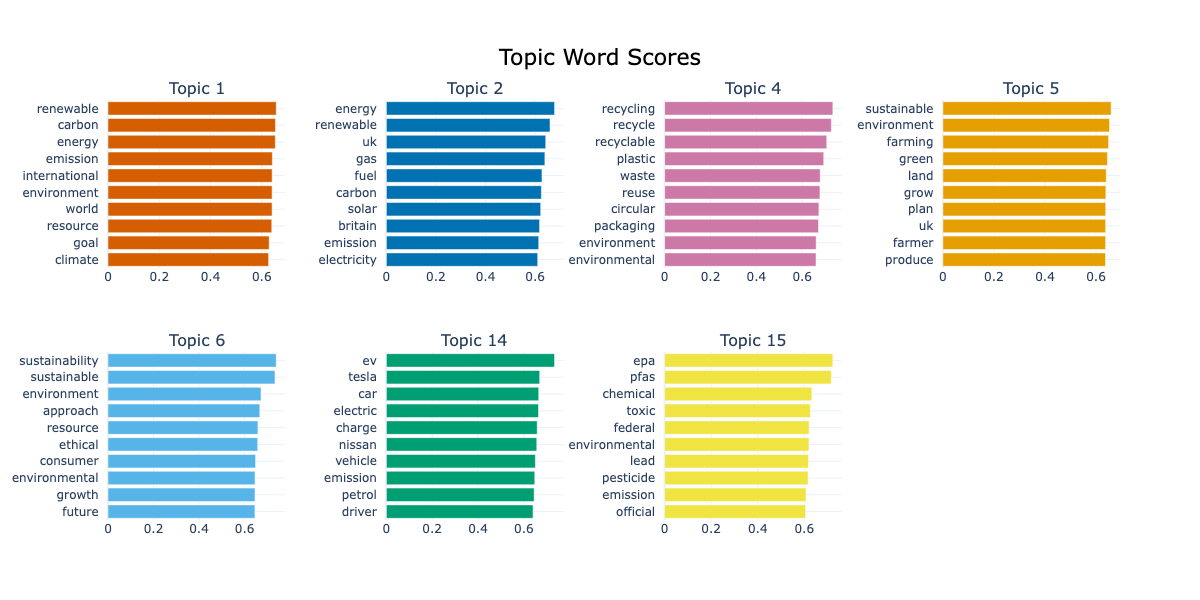}
    \caption{Bar chart representation of topic modelling results using the BERTopic from The Guardian dataset.}
    \label{fig:bar_guardian}
\end{figure}

As illustrated in Figure \ref{fig:bar_guardian}, we have selected seven topics from the BERTopic model applied to The Guardian dataset for analysis. The bar chart displays the scoring (c-TF-IDF score) of individual words (terms) within each topic, ranging from 0 to 1. Higher scores indicate a greater significance of the word in its respective topic, reflecting its substantial influence on the topic's importance.

It is evident that under optimal parameter tuning, the BERTopic model yields scores of 0.6 or higher for every word within the seven selected topics. Topics 1 and 5 demonstrate the most uniform distribution of word scores among the ten words analysed, indicating a lack of distinct representative words that characterise these topics compared to others. The majority of words in Topic 1 pertain to global aspects of energy and the environment, with `renewable' achieving the highest score and `climate' the lowest. In contrast, Topic 5 does not feature words related to `world'; instead, it includes terms such as `uk' and `farmer', focusing more on national-level sustainable agricultural development. 

The remaining five topics each feature one or two words with significantly higher scores. For instance, in Topic 14, the score for `ev' is notably higher than that of the other words. This is indicative of the strength of BERTopic as a language model that utilises contextual analysis to effectively summarise and comprehend abbreviations like `ev'. Other related terms, such as brands including `Tesla' and `Nissan', aid the model in more swiftly grasping the significance of `ev'. The descending order of scores also illustrates the prioritisation and segmentation of these words within the topic.

We acknowledge that BERTopic performs admirably in interpreting text within The Guardian dataset, aggregating relevant vocabulary under a unified topic. However, an analysis of the highest-scoring words from Topics 4 and 6 reveals that words of different grammatical forms often convey essentially the same meaning. Both `recycling' and `recycle' in the text reflect the concept of resource recovery, while `sustainability' and `sustainable' repeatedly express the notion of environmental sustainability. We believe that by leveraging the strengths of language models, including part-of-speech normalisation and selective word comparison, BERTopic can offer researchers a more comprehensive vocabulary pertaining to each topic.

\begin{figure}
    \centering
    \includegraphics[scale=0.4]{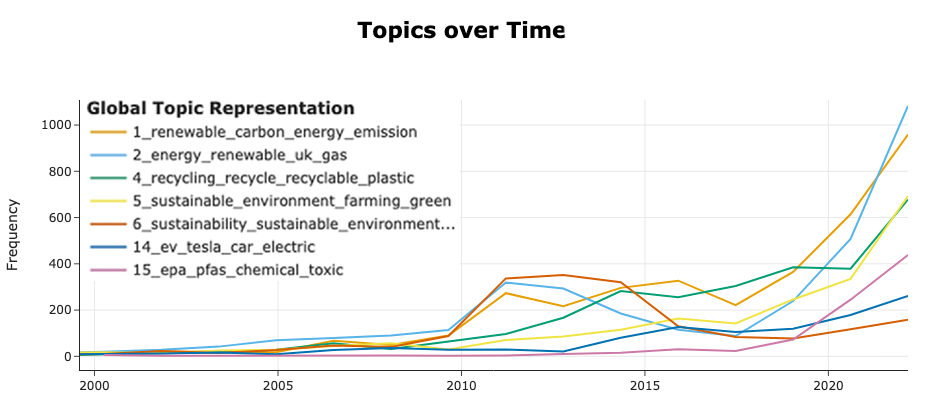}
    \caption{This line chart shows the frequency of occurrence of seven topics generated by the BERTopic model trained on The Guardian dataset (1999 to 2023).}
    \label{fig:dynamic_guardian}
\end{figure}

The frequencies and corresponding times of the seven topics are presented in Figure \ref{fig:dynamic_guardian}. Topic 1 shows a growing trajectory, rising significantly starting around 2017. This upgrade is timed to coincide with key international developments such as the 2016 Paris Agreement, highlighting the growing focus in the UK news on the synergies between renewable energy adoption and carbon emissions. Topic 15 experienced a surge in frequency around 2019. The topic continues to gain visibility, and this rise may be related to increased public concern about chemical pollution incidents and environmental health. The increase in frequencies of all topics post-2015 suggests an overall growing public and journalistic interest in environmental issues. This period seems to be pivotal in shaping public attention to sustainability and environmental topics.

\begin{figure}
    \centering
    \includegraphics[scale=0.4]{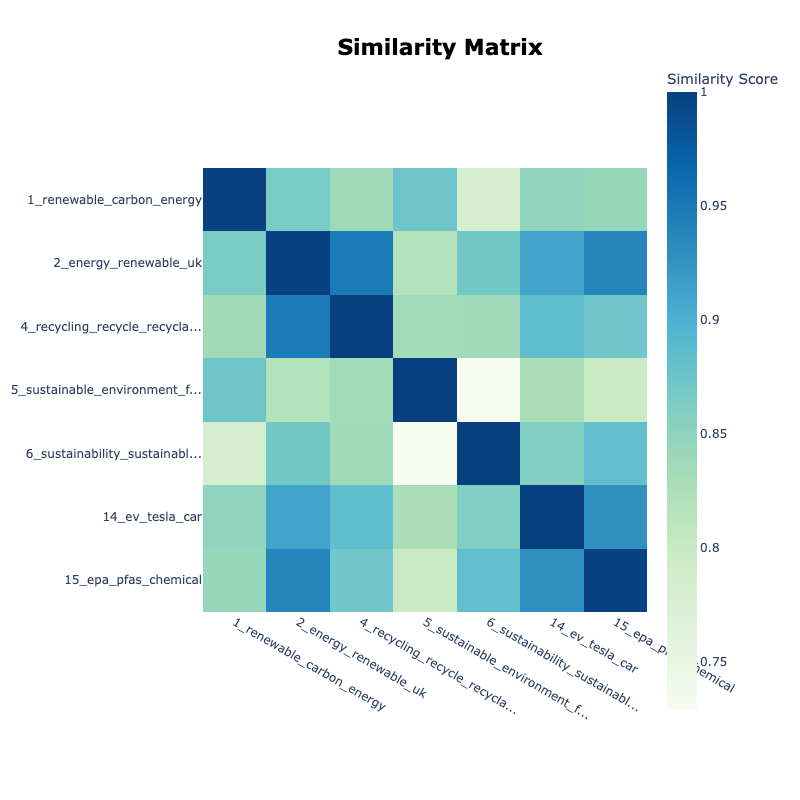}
    \caption{Heatmap of the similarity matrix between topics from The Guardian dataset using the BERTopic.}
    \label{fig:heatmap_guardian}
\end{figure}

Figure \ref{fig:heatmap_guardian} presents a heatmap of the similarity matrix for topics generated by the BERTopic model on The Guardian dataset. These seven topics correspond to the ones shown in Figure \ref{fig:bar_guardian}. Deeper shades represent higher cosine similarity, whereas lighter shades indicate lower similarity. Since we get more topics in the Guardian dataset than in the Reddit and Twitter datasets, we believe the heatmap in Figure \ref{fig:heatmap_guardian} facilitates a comparative analysis of the interrelationships among the topics.

The relationship between Topic 2 and Topic 15 is indicated by the darker shades on the heatmap, suggesting a strong connection. Further investigation into the vocabulary reveals that discussions on renewable energy in The Guardian are considered within a broader context, where the impact of chemical regulations and substances such as PFAS is indispensable. There could be a policy-oriented discourse that integrates renewable energy initiatives with the regulation of hazardous chemicals. Compared to its similarity with Topic 15, Topic 2 may have a closer relationship with Topic 4. Topic 2 and Topic 4 both involve environmental sustainability. `Recycling' is a key element in Topic 4 and an integral part of sustainable practices. The discussions in Topic 4 likely address practical issues of environmental sustainability, such as `waste' and `packaging' which complement the broader topic of sustainability in Topic 2.

The relatively superficial relationship between Topics 5 and Topic 6 is particularly intriguing, given that both seemingly focus on sustainability and the environment. However, the low similarity score indicates that each topic explores different facets of this broad thematic focus. Topic 5 encompasses terms such as `sustainable', `environment', `uk', `farmer', and `produce'. These terms collectively suggest discussions centred around sustainable agriculture and production. `Farmer', `farming', and `produce' highlight the practical implementation of sustainable practices within agricultural activities under the CE. The presence of `green' and `land' signifies concerns over ecological considerations in land use. The terms `plan' and `uk' may denote specific strategies within the British agricultural sector. Conversely, Topic 6 is characterised by words like `sustainability', `sustainable', `approach', `resource', `ethical', `growth', and `future'. While both topics emphasise sustainability and the environment, the additional presence of terms such as `approach', `resource', and `ethical' indicates a broader conceptual discourse. The term `ethical' implies discussions on the moral imperatives of sustainability, potentially involving corporate responsibility, while `consumer' shifts the focus to individuals and the market. `growth' and `future' suggest a forward-looking perspective, contemplating the long-term trajectory of sustainable development. Although there is thematic overlap in the general concern for sustainability and environmental issues, BERTopic effectively reveals the specific contexts in which these terms are used, highlighting divergences in discourse. Topic 5 focuses more on the practicalities of sustainable agriculture and environmental management. In contrast, Topic 6 appears to discuss sustainability from a broader perspective, encompassing ethical considerations, consumer habits, and approaches aimed at promoting long-term sustainable growth.

Further inspection of the heatmap we obtained, we found topic 14 and topic 15, which have less overlap with other topics. The lighter shading indicates this visually, suggesting that the discourse within these topics is more isolated and specialised. Topic 14’s focus on electric vehicles also suggests that the Guardian’s discussion, largely represented by Tesla, opens up a dialogue with the wider environment. Likewise, Topic 15 deals with chemical regulations, with a possible emphasis on PFAS and EPA oversight, suggesting that the focus of the discussion is on the specificity of regulation and science and does not broadly intersect with the rest of the topic landscape. Despite the differing focuses of Topics 14 and Topic 15, we observed that their substantial similarity may stem from a shared concern for environmental impact and regulation. Both topics likely engage in a broader narrative about reducing harmful emissions, whether through the promotion of electric vehicles or the regulation of toxic chemicals. Furthermore, both topics appear to involve human activities (driving in Topic 14 and chemical usage in Topic 15) and governmental measures to mitigate associated risks. This convergence may reflect an increasing societal and regulatory discourse focused on sustainability and mitigating environmental hazards.

\begin{figure}
    \centering
    \includegraphics[scale=0.4]{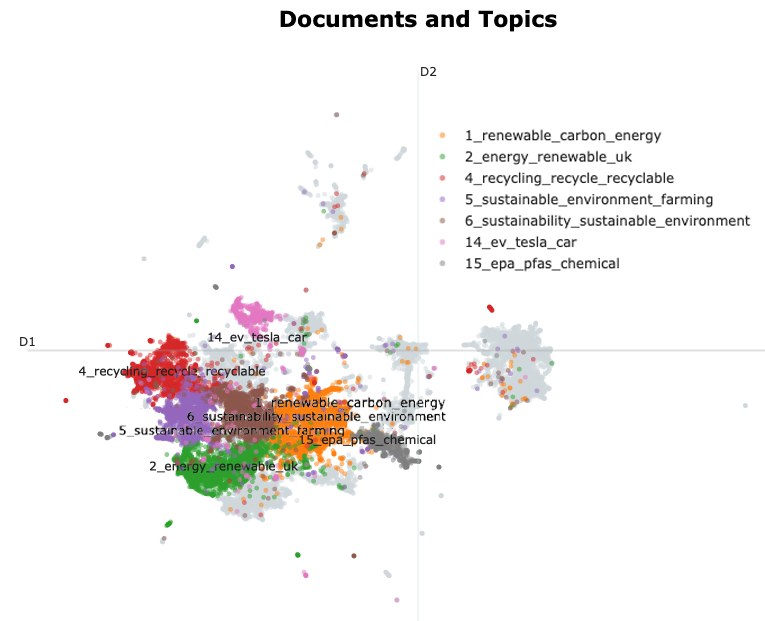}
    \caption{Scatter plot visualisation of documents and their associated topics from The Guardian dataset, clustered by similarity via the BERTopic model.}
    \label{fig:scatter_guardian}
\end{figure}

Figure \ref{fig:scatter_guardian} provides a two-dimensional representation of the topics generated under BERTopic for The Guardian dataset, with each data point representing an individual document. The position of each document in two-dimensional space (depicted by D1 and D2) reflects the topic of the document as determined by BERTopic's semantic processing. A colour-coding scheme is employed to demarcate document clusters that correspond to the discrete topics unearthed by BERTopic. 

According to this scatter plot, we can notice that the cluster of Topic 14 (pink colour) is somewhat separated from the rest of the topic clusters. This means that the news around electric vehicles in The Guardian is more of a dedicated discussion to this area, focusing on key industry players such as Tesla. While this topic may not be at the centre of most discussions in The Guardian news, it can be a good representation of the impact of innovation trends, and CE being carried out in transport. Additionally, although Topic 15 (grey colour) is similar to other clusters, it has a smaller size which may reflect the specificity of discussions around toxic chemicals. The size of clusters in this topic shows there is a specialised and smaller number of news about chemical terms and EPA in The Guardian dataset.

\begin{figure}
    \centering
    \includegraphics[scale=0.7]{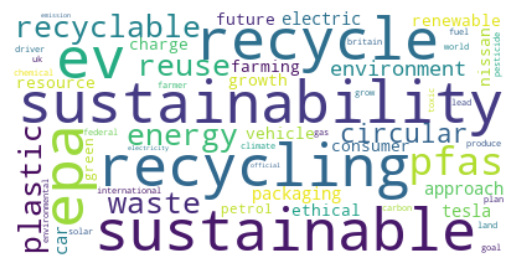}
    \caption{Word cloud visualising the prominence of terms in topics derived from The Guardian dataset using the BERTopic model.}
    \label{fig:wc_guardian}
\end{figure}

In the field of topic modelling, word clouds serve as a visual heuristic method, displaying terms that occupy central nodes in the discourse network \cite{kalmukov2021using}.  Figure \ref{fig:wc_guardian} shows the topic vocabulary graph cloud generated by BERTopic on The Guardian dataset. The visual prominence of terms such as `sustainability', `sustainable', `recycling', and `energy' within the cloud is indicative of their pivotal role in the corpus, highlighting the dataset's focus on environmental conservation, renewable energy, and the principles of the CE. These terms underscoring the centrality of sustainable energy in CE are captured within The Guardian news. Surrounding these core terms are secondary and substantial, concepts such as `ev', `electric', and `vehicle', which given their relative visual weight (size), suggest substantive topics that concern technological advancements in transportation and the transition. Moreover, the adjacency of concepts such as `ethical', `consumer', `growth', and `future' alongside 'sustainability' within the visualisation elucidates an interplay of notions.

\subsubsection{MOO} \label{sec:theguardian_moo}

The topics generated by BERTopic on The Guardian dataset, optimised via MOO, provide a more nuanced view of the various discussions related to the CE and other significant topics because the multi-objective optimisation balances coherence, diversity, and perplexity. This balance allows the model to capture a broader spectrum of topics, uncovering subtle and diverse topics that might be overlooked when focusing on a single objective. By optimising multiple objectives simultaneously, we enable the discovery of richer and more comprehensive topic representations within the data. Notably, Topic 2, which includes terms such as `renewable', `energy' and `uk' focuses heavily on UK environmental and energy-related issues. This aligns well with the core concerns of the CE, showcasing how MOO helps capture a broader and more diverse range of topics compared to single-objective optimisation. The inclusion of relevant terms such as `solar' and `tory' suggests an intersection of political and environmental discussions, providing a richer context for researchers examining the policy implications of climate change.

Another critical topic, Topic 4, brings together terms like `sustainability', `resource' and `future' directly relating to the CE. This topic reflects broader discussions about sustainable development goals and the future of resource management, emphasising the model's ability to capture key aspects of CE debates. By incorporating terms such as `deforestation' and `energy', MOO allows the model to go beyond narrow environmental concerns, offering insights into the strategic planning and forward-looking goals of CE.

The diversity of topics presented in \Cref{tab:moo_topics_guardian} also reveals insights into current social and political affairs, such as Topics 1 and 3, which focus on UK politics, and Topic 11, which highlights the impact of the COVID-19 pandemic. These topics provide a wider socio-political context, reflecting the broader scope of media coverage in The Guardian. Overall, MOO enables a more comprehensive extraction of topics, providing researchers with a broader spectrum of relevant topics related to the CE and beyond, ensuring that key dimensions of current societal challenges are well-represented in the analysis.

On The Guardian platform, the results of SOO and MOO highlight different optimisation focuses. SOO primarily maximises topic coherence, focusing on the semantic consistency of the generated topics. Topic models optimised by SOO, such as BERTopic, show relevance in the generated topic words, with terms like `energy', `gas', and `emission' closely tied to environmental sustainability. However, due to its sole focus on coherence, the number of generated topics is small (e.g., 2 topics), resulting in limited diversity.

In contrast, MOO generates a more diverse set of topics by optimising coherence, diversity, and perplexity simultaneously. The BERTopic model in MOO produces 20 topics, covering a wider range of circular economy topics, such as `recycling', `renewable energy', and `sustainability'. This demonstrates that the MOO method maintains a high level of semantic coherence while enhancing topic diversity. Moreover, the lower perplexity in MOO indicates that the model is more stable when predicting unseen text. Overall, the MOO model increases topic diversity and model robustness while maintaining high coherence, making it suitable for analysing complex text data.

\begin{table}
\centering
\caption{The topic terms obtained from the BERTopic model with the best hyperparameter values on The Guardian dataset (MOO).}
\resizebox{\textwidth}{!}{
\begin{tabular}{c|l}
\bottomrule
\multicolumn{1}{c}{Topic} & \multicolumn{1}{c}{Topic Term} \\ \hline
0 & australia, australian, nsw, bst, queensland, sydney, gmt, labor, election, people \\ 
1 & corbyn, tory, mp, uk, brexit, conservative, eu, election, britain, question \\ 
2 & renewable, energy, tory, uk, britain, solar, consumer, carbon, conservative, emission \\ 
3 & cop27, cop26, carbon, emission, world, global, energy, international, 20230, renewable \\ 
4 & sustainability, sustainable, resource, future, create, ethical, develop, challenge, organisation, environment \\ 
5 & rio, world, sdgs, earth, resource, environment, international, global, deforestation \\ 
6 & meat, food, farming, feed, farmer, beef, produce, dairy, crop, agriculture, farm \\ 
7 & car, emission, vehicle, petrol, drive, electric, charge, diesel, driver, carbon \\ 
8 & recycling, recycle, recyclable, rubbish, waste, plastic, landfill, reuse, circular, litter \\ 
9 & feel, moment, need, want, share, use, think, tell, say, medium \\ 
10 & candidate, trump, biden, voter, democrat, election, official, senator, democrats, republicans \\ 
11 & covid, coronavirus, vaccine, vaccination, lockdown, world, gmt, outbreak, pandemic, die \\ 
12 & ftse, inflation, eurozone, fed, uk, gmt, market, investor, gdp, trading \\ 
13 & arsenal, player, sport, liverpool, club, league, final, chelsea, goal, team \\ 
14 & housing, property, house, landlord, building, site, tenant, land, build, resident \\ 
15 & clothe, wear, garment, fashion, clothing, textile, retailer, consumer, shirt, fabric \\ 
16 & epa, pfas, chemical, toxic, pesticide, environmental, federal, emission, pollution, lead \\ 
17 & ukraine, ukrainian, russia, putin, kyiv, kremlin, eu, world, russian, attack \\ 
18 & golf, tournament, final, putt, forehand, mcilroy, tee, open, player, backhand \\ 
19 & samsung, smartphone, tablet, android, device, price, app, screen, ipad, fold \\
\bottomrule
\end{tabular}}
\label{tab:moo_topics_guardian}
\end{table}

\subsection{Reddit}

\subsubsection{SOO} \label{sec:reddit}

To explore discussions related to the circular economy, we utilised the Reddit API to collect relevant posts for our topic modelling. The Reddit Official API facilitates the extraction of posts by subreddit (for a particular topic) or keyword searches. We specifically targeted subreddits related to the circular economy and filtered for posts containing the keyword `circular economy' across broader discussions. This approach enabled us to compile thousands of posts, including their titles, contents, and time. We store all data in JSON files to ensure that the data format is consistent across platforms.

The BERTopic model (Table \ref{tab:besthpsbertreddit}) identifies two topics. The first topic integrates core aspects of the circular economy such as `recycling', `sustainable' and `waste' pointing towards operational and environmental concerns. The second topic seems to focus on the business and ideational perspectives, including terms like `business model' and `circularity', which suggest strategic to circular economy practices. BERTopic provides a better understanding of the contextual relationships between words compared to traditional LDA models because of the presence of the language model BERT. By utilising BERT's deep semantic understanding, it can link concepts such as environment and sustainability, generating related words such as `circular economy' and `sustainable'. The word embedding of the BERT model allows the model to capture the link between specific activities such as `recycling' and `plastic waste' and the broader topic of circular economy.

The CorEx model (Table \ref{tab:besthpscorexreddit}) results show a diverse set of subtopics over three topics. The first topic is focused on fundamental actions and strategies like `reduce' and `consumption'. The second topic is more oriented towards community and engagement with terms like `help', `thank' and `people'. The third topic captures industrial and infrastructural elements, with terms such as `material', `battery' and `recycling' reflecting specific operational sectors. The best hyperparameter results are more comprehensive than BERTopic because CorEx focuses more on discovering the greatest common information between variables, not just combinations of words that occur frequently in LDA. The anchor vocabulary of \textbf{3R Rules} that we set up in our experiments to guide the model to focus more on specific thematic directions also helped CorEx to identify non-intuitive topics in Reddit. For example, Topic 3 shows that words such as `material', `battery' and `recycling' reflect industry-specific discussions. We believe this provides valuable insight for industry policymakers.

Finally, the LDA model (Table \ref{tab:besthpsldareddit}) produces two topics that, while similar, emphasise different facets of circular economy discussions. Both topics feature `circular', `economy' and `waste' as central terms but differ slightly in focus. The first topic includes `new' and `need', which may indicate discussions around solutions within the circular economy. The second topic includes `business' and `good' likely reflecting on the efficacy and benefits of circular economy practices in business contexts. From the words frequently appearing in the two topics, such as `economy', `circular', `waste' and `sustainable', it can be seen that the LDA model directly captures the core words related to circular economy. While this approach may not be as sensitive as deep learning-based BERTopic, it has a unique advantage in providing intuitive topic distribution.

\begin{table}
\centering
\caption{The topic terms over 2 topics obtained from the BERTopic model with the best hyperparameter values on the Reddit dataset.}
\resizebox{\textwidth}{!}{
\begin{tabular}{c|l}
\bottomrule
\multicolumn{1}{c}{Topic} & \multicolumn{1}{c}{Topic Term}                                                                                            \\ 
\hline
0                         & circular\_economy, recycling, recycle, plastic waste, circular, sustainable, reuse, environment, sustainability, waste     \\
1                         & circular\_economy, circular, business\_model, idea, sustainable, recycling, sustainability, resource, circularity, recycle  \\
\bottomrule
\end{tabular}}
\label{tab:besthpsbertreddit}
\end{table}

\begin{table}
\centering
\caption{The topic terms over 3 topics obtained from the CorEx model with the best hyperparameter values on the Reddit dataset.}
\resizebox{\textwidth}{!}{
\begin{tabular}{c|l}
\bottomrule
\multicolumn{1}{c}{Topic} & \multicolumn{1}{c}{Topic Term}                                                      \\ 
\hline
0                         & reduce, consumption, waste, world, enable, city, use, base, produce, bag            \\
1                         & help, thank, work, project, reuse, hi, look, people, like, currently                \\
2                         & recycle, material, battery, company, loop, london, recycling, solar, public, clean  \\
\bottomrule
\end{tabular}}
\label{tab:besthpscorexreddit}
\end{table}

\begin{table}
\centering
\caption{The topic terms over 2 topics obtained from the LDA model with the best hyperparameter values on the Reddit dataset.}
\resizebox{\textwidth}{!}{
\begin{tabular}{c|l}
\bottomrule
\multicolumn{1}{c}{Topic} & \multicolumn{1}{c}{Topic Term}                                                      \\ 
\hline
0                         & economy, circular, new, waste, product, plastic, help, recycle, sustainable, need   \\
1                         & circular, economy, waste, plastic, business, sustainable, product, new, good, help  \\
\bottomrule
\end{tabular}}
\label{tab:besthpsldareddit}
\end{table}

\paragraph{Visualisation and Analysis}\label{sec:professionals}

\begin{figure}
    \centering
    \includegraphics[scale=0.5]{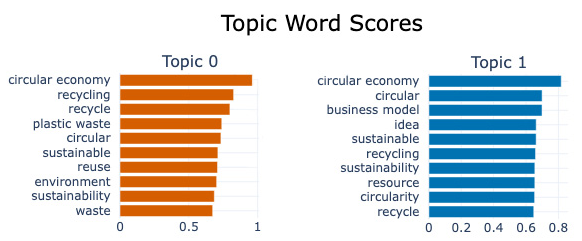}
    \caption{Bar chart representation of topic modelling results using the BERTopic from Reddit dataset.}
    \label{fig:bar_reddit}
\end{figure}

Figure \ref{fig:bar_reddit} illustrates all topics generated by the BERTopic model on the Reddit dataset using the best hyperparameters. Through hyperparameter optimisation, we determined that configuring the generation of two topics yields optimal results on the Reddit dataset. Hence, the diagram displays only these two topics alongside the scoring of various words. The word scores of Figure \ref{fig:bar_reddit} (0 to 1), adhering to the same criteria as illustrated in Figure \ref{fig:bar_guardian}, serve as an indicator of the contextual relevance of words within a topic. This provides an empirical foundation for the analysis of topics generated by the BERTopic model on the Reddit dataset.

Unlike the results from The Guardian dataset, the topics generated on Reddit by the BERTopic model include not only individual words but also phrases composed of two words. In Topic 0, terms such as `circular economy', `recycling' and `plastic waste' dominate, indicating that the overall narrative focuses on environmental management, particularly within the scope of waste management. The recurrent topic of `circular economy' supports discussions that emphasise material reuse as a strategy for reducing waste production. 

Conversely, Topic 1 encapsulates vocabularies that link environmental issues with business. Words such as `business model', `sustainable' and `resource circularity' substantiate these linkage. Topic 1 points towards the concept of sustainable development, which is integral to business innovation, combining economic incentives with ecological responsibilities. The repeated occurrence of words like `circular economy' and `recycling' across both topics underscores their central role in the discourse. However, it can be observed in Topic 1 which is generated by the BERTopic model on the Reddit dataset, the emphasis on `business model' and `idea' suggests a more strategic perspective than the operational viewpoint found in Topic 0.

\begin{figure}
    \centering
    \includegraphics[scale=0.3]{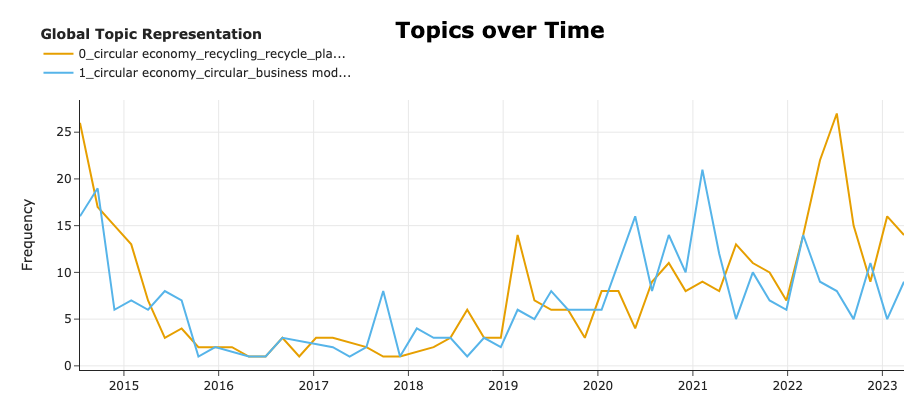}
    \caption{This line chart shows the frequency of occurrence of the two Topics generated by the BERTopic model trained from the Reddit dataset in recent years.}
    \label{fig:dynamic_reddit}
\end{figure}

In Figure \ref{fig:dynamic_reddit}, Topic 0 involves terms such as `circular economy', `recycling', `recycle', `plastic waste', and related environmental terms. After 2015, it shows a declining trend, stabilising in subsequent years with fluctuations but a continuous presence. This trend may indicate that discussion peaks on Reddit could have been triggered by specific events or policies. Each fluctuation could reflect the evolving interest in the circular economy (CE) on Reddit. As concepts of waste management and sustainable practices increasingly penetrate collective consciousness, they might also represent an evolving discourse. Topic 1 (yellow line) focuses on `circular economy', `circular', `business model', and `idea'. Topic 1 displays greater volatility and pronounced peaks, particularly in 2022, which may relate to significant international decisions on climate and economy or the easing of the COVID-19 pandemic. Topic 1 shows more distinct peak trends, while Topic 0 exhibits more consistency after an initial decline. This may suggest that discussions on Reddit about circular economy business models are more sensitive to specific events and also reflect ongoing discussions within the community about the development of CE.

\begin{figure}
    \centering
    \includegraphics[scale=0.35]{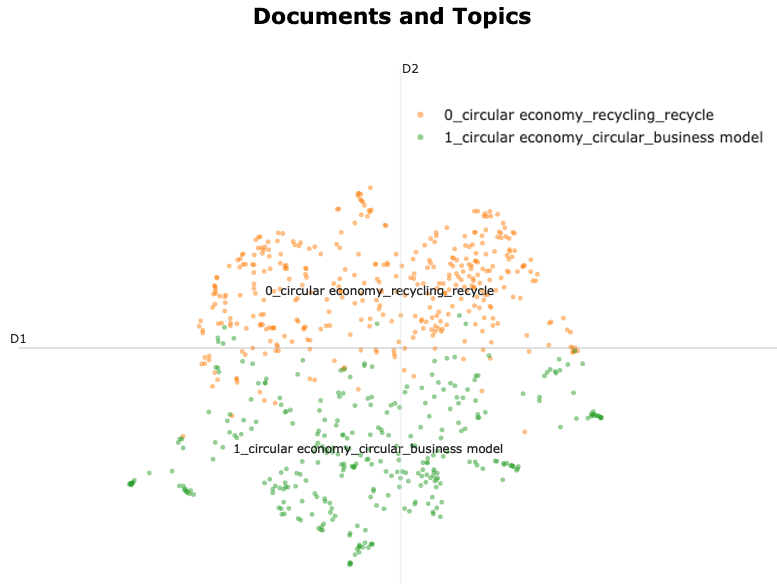}
    \caption{Scatter plot visualisation of documents and their associated topics from Reddit dataset, clustered by similarity via the BERTopic model.}
    \label{fig:scatter_reddit}
\end{figure}

Figure \ref{fig:scatter_reddit} reveals the document distribution and description of two topics as a scatter plot. Topic 0 is shown in orange and summarises the discussion with `circular economy' and `recycling'. Compared with the distribution of Topic 1, it can be found that the data points are relatively denser. It shows that the conversation space on Reddit is more concentrated, and users’ conversations on Reddit are more focused on the views between CE and recycling. The green Topic 1 discussion revolves around `circular economy' and `business model'. It shows that the conversation space on Reddit is wider, and users can have a wider discussion about integrating sustainability into business models. Although these two clusters have different topic focuses, there are still some data points that are close to each other and another topic cluster, reflecting a certain dependence between recycling and business models.

\begin{figure}
    \centering
    \includegraphics[scale=0.7]{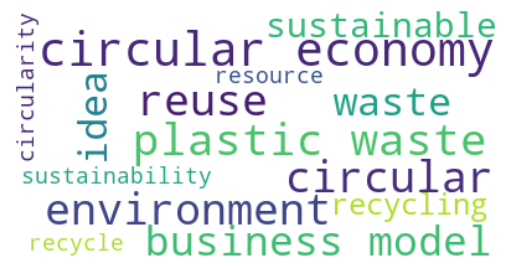}
    \caption{Word cloud visualisation of the prominence of terms in topics derived from Reddit dataset using the BERTopic model.}
    \label{fig:wc_reddit}
\end{figure}

Figure \ref{fig:wc_reddit} is a word cloud produced by BERTopic on the Reddit dataset, which presents the main topic words of CE in different sizes and positions. Unlike the word cloud (Figure \ref{fig:wc_guardian}) from The Guardian dataset, `circular economy' appears prominently in the word cloud, its large size indicating a central role in the discourse. Adjacent to `circular economy' are terms such as `reuse' and `resource'. Although these terms are not as visually prominent as `circular economy', they remain crucial in the narrative, indicating strong discussions about re-usable resources. The central appearance of `plastic waste' marks focused conversations around plastic consumption and waste management. The visibility of this term in the cloud summarises community engagement with specific environmental impacts, resonating with global concerns over plastic pollution. Intersecting with these environmental terms, `business model' anchors the discourse in the economic domain. The combination of `business model' and `circular economy' reflects discussions on business structures aligned with circularity and resource conservation.

The word cloud (Figure \ref{fig:wc_reddit}) not only captures the direction of Reddit user's discussions about CE but also provides insight into the collective focal points of the community, thereby reflecting the broader public engagement and attention towards sustainability.

\subsubsection{MOO} \label{sec:reddit_moo}

As shown in the \Cref{tab:moo_topics_reddit}, topic 1 includes terms such as `circular', `economy' and `environment' demonstrating the model's ability to capture the broader intersection between environmental issues and economic strategies. This reflects the community-driven nature of Reddit, where discussions often highlight the socio-economic implications of environmental policies. The inclusion of words like `economic' and `environment' suggests a nuanced understanding of the relationship between sustainable practices and economic outcomes.

Topic 3 focuses on recycling, with terms such as `recycling', `reuse' and `reusable'. Topic 3 highlights the core components of the CE. This shows how MOO effectively captures the practical aspects of waste management and resource reuse, which are crucial topics in any CE discourse. By extending beyond simple coherence optimisation, MOO reveals the depth of user-generated content on Reddit, which often centres on community-led initiatives and grassroots movements.

The diversity of topics in \Cref{tab:moo_topics_reddit} demonstrates how MOO can uncover not only the technical aspects of CE but also the societal and economic dimensions. For instance, Topic 2 brings together terms like `eu', `circular' and `sustainable', indicating ongoing discussions about European sustainability and resource management policies. Overall, the MOO results provide a well-rounded analysis of CE-related discussions, enabling researchers to better understand the key issues being debated on platforms like Reddit.

On the Reddit platform, the topic model generated by SOO optimisation performs poorly in terms of coherence (negative C NPMI value), likely due to the scattered and informal nature of user discussions on Reddit. The SOO results show a small number of topics (e.g., 2), which may not fully capture the breadth of public discussion on the platform. Additionally, SOO’s topic words contain more redundant information, such as `say' and `year', reflecting SOO’s limitations when dealing with dispersed data.

MOO optimisation, on the other hand, generates more topics (14) and significantly improves the diversity score. This suggests that, despite the scattered nature of user discussions on Reddit, MOO can still capture a more diverse range of topic structures. Furthermore, although MOO’s consistency score is lower, the improvement in perplexity indicates that the model performs better when dealing with unstructured and varied content. Therefore, MOO is better at balancing consistency and diversity on Reddit, allowing it to capture more of the platform's public concerns.

\begin{table}
\centering
\caption{The topic terms obtained from the BERTopic model with the best hyperparameter values on the Reddit dataset (MOO).}
\resizebox{\textwidth}{!}{
\begin{tabular}{c|l}
\bottomrule
\multicolumn{1}{c}{Topic} & \multicolumn{1}{c}{Topic Term} \\ \hline
0  & sustainability, sustainable, lca, questionnaire, circular, survey, feedback, project, idea, question \\ 
1  & circular, economy, environment, economic, sustainability, emission, future, energy, resource, innovation \\ 
2  & eu, circular, european, sustainable, europe, 2030, ubi, environment, consumer, project \\ 
3  & recycling, recycle, reuse, reusable, circular, sustainable, upcycled, idea, use, packaging \\ 
4  & recycling, recycle, recyclate, plastic, microplastic, environmental, circular, packaging, sustainable, waste \\ 
5  & podcast, circular, sustainable, sustainability, resource, circularity, consumption, lecture, opportunity, innovation \\ 
6  & circular, clothing, outfit, clothe, garment, fashion, textile, survey, wear, feedback \\ 
7  & recycling, environment, recycle, circular, energy, greenwashing, sustain, cyclical, environmentalist, ipcc \\ 
8  & circular, circularity, opendesign, modeindustrie, design, video, eco, materiali, biomimicry, animation \\ 
9  & repair, repairable, fix, repairer, repairability, gadget, ifixit, device, electronic, check \\ 
10 & recycle, recycling, recycled, recyclemore, renewable, zerowaste, circular, reuse, eco, waste \\ 
11 & recycle, recycling, recyclable, ev, solar, pv, energy, electrocatalyst, battery, circular \\ 
12 & webinar, circular, upcoming, register, coursera, hub, registration, event, workshop, course \\ 
13 & macarthur, circular, ellen, toolkit, report, programme, foundation, interested, 2020, project \\ 
\bottomrule
\end{tabular}}
\label{tab:moo_topics_reddit}
\end{table}

\subsection{Twitter}

\subsubsection{SOO} \label{sec:twitter}

While exploring user attention on the CE from Twitter, we utilised Twitter’s official API to collect posts containing the hashtag `Circular Economy'. However, starting in 2023, Twitter's official API began restricting the scope of user searches, limiting us to accessing data from the past year. We conducted data cleaning on all collected data, retaining core information such as the content and timing of tweets. We also removed duplicates, replies, and reposted tweets, ultimately preserving 3,922 English tweets for our model learning.

Table \ref{tab:besthpsberttwitter} shows topics from the Twitter dataset by the BERTopic model. Topic 0 clusters around the principles of the circular economy, emphasising the integration of recycling and waste management into sustainable practices. The convergence of `sustainable' and `circular' within this topic highlights a prevailing narrative focused on the implementation of resource-efficient economies. Topic 1 projects a future-oriented discussion, where the aspiration for a `sustainable future' is intricately linked to the notion of `circularity', reflecting the community's advocacy for systemic change towards enduring environmental stewardship.

Furthermore, Topic 2 focuses on the recyclability of packaging materials, illustrating a deepened public concern over the lifecycle of consumables and the pursuit of sustainable resource utilisation. In an interesting thematic diversion, Topic 3 intertwines charity-related activities with the circular economy, suggesting that philanthropic efforts are increasingly oriented towards supporting sustainable practices. Topics 5 through 9 pivot to a technological perspective, shedding light on discussions around the resale and recycling of network equipment. The presence of terms specific to networking hardware such as `router' and `Cisco' points to a vibrant discourse on the CE within the tech sector.

\begin{table}
\caption{The topic terms over 10 topics obtained from the BERTopic model with the best hyperparameter values on the Twitter dataset.}
\resizebox{\textwidth}{!}{
\begin{tabular}{c|l}
\bottomrule
\multicolumn{1}{c}{Topic} & \multicolumn{1}{c}{Topic Term}                                                                                                                                                                                                                                                                                                 \\ 
\hline
0     & \begin{tabular}[c]{@{}l@{}}circular economy, sustainability circulareconomy, circulareconomy sustainability, recycling,circulareconomy, recycle,wastemanagement, circular, sustainable,\\sustainability,\end{tabular}                                                                                                                \\
1     & \begin{tabular}[c]{@{}l@{}}circulareconomy, sustainability circulareconomy, circulareconomy sustainability, circulareconomy circulareconomy, circular economy, sustainable future, circularity,\\circular, create sustainable, sustainable,\end{tabular}                                                                               \\
2     & \begin{tabular}[c]{@{}l@{}}sustainablepackage recyclability,recyclability package,circulareconomy recycledmaterial,sustainablepackage,recycledmaterial resourceefficiency,\\package sustainability,recycle sustainability,sustainability circulareconomy,circulareconomy sustainability,circulareconomy recycle,\end{tabular}  \\
3     & \begin{tabular}[c]{@{}l@{}}monday charitynee,supportlocal mondaythought,charity need,need charity,charity circulareconomy,work charity,support need,need monday,support amazing,\\help support,\end{tabular}                                                                                                                   \\
4     & \begin{tabular}[c]{@{}l@{}}ban plastic,plasticwaste export,circulareconomy endwastecolonialism,plastic waste,endwastecolonialism tell,wasteshipment breakfreefromplastic,\\mess wastetrade,waste export,petition wastetrade,wastetrade,\end{tabular}                                                                           \\
5     & router processor,processor cisco,cisco 12000,performace router,router cisco,forsale cisco,cisco circulareconomy,2951 router,cisco 2921,asr1001 router,                                                                                                                                                                         \\
6     & forsale cisco,c3650 48td,ws c3650,c3650,switch 3650,48 port,cisco,cisco ws,cisco circulareconomy,48td catalyst,                                                                                                                                                                                                                \\
7     & 10xge port,adapter xfp,xfp port,spa 1xtenge,cisco spa,port adapter,gigabit ethernet,port multifunction,forsale cisco,1xtenge xfp,                                                                                                                                                                                              \\
8     & extreme network,24pt poe,switch extreme,summit x450e,summit x460,summit x450a,poe switch,cable extreme,x450e 24p,x460 24,                                                                                                                                                                                                      \\
9    & roast cup,roaster chef,roaster,coffee roaster,mb roast,roast,roaster barista,contemporary roaster,cheflife chefsofinstagram,cup chef,                                                                                                                                                                                          \\
\bottomrule
\end{tabular}}
\label{tab:besthpsberttwitter}
\end{table}

As show in Table \ref{tab:besthpscorextwitter}, the CorEx model's analysis of the Twitter dataset reveals a nuanced landscape of sustainability discourse. Topic 0 focuses on reducing environmental impacts, emphasising emissions, sustainable resource usage, and the need for carbon footprint minimization. Topic 1 reflects community engagement in sustainability through reuse and repair, underscored by charitable activities. Topic 2's dialogue centres on recycling specifics, indicating targeted discussions on material recovery and recycling industries. Topic 3 diverges to a commercial domain, highlighting events and technology sales, possibly in an ecological context. Finally, Topic 4 focused on global waste management policies. 

\begin{table}
\centering
\caption{The topic terms over 5 topics obtained from the CorEx model with the best hyperparameter values on the Twitter dataset.}
\resizebox{\textwidth}{!}{
\begin{tabular}{c|l}
\bottomrule
\multicolumn{1}{c}{Topic} & \multicolumn{1}{c}{Topic Term}                                                                                           \\ 
\hline
0                         & reduce, emission, carbon, resource, waste, sustainable, planet, recyclability, sustainablepackage, co2                   \\
1                         & reuse, charity, item, repair, charitynee, monday, supportlocal, amazing, mondaythought, volunteer                        \\
2                         & recycle, recycling, tyre, tyrerecycle, battery, tirerecycle, tyrerecovery, recyclingbusiness, noexcusesgocircular, tire  \\
3                         & forsale, register, event, cisco, april, conference, 2023, port, summit, join                                             \\
4                         & breakfreefromplastic, wastetrade, baselconvention, wasteshipment, ban, endwastecolonialism, export, mess, tidy, eu       \\
\bottomrule
\end{tabular}}
\label{tab:besthpscorextwitter}
\end{table}

Finally, according to the topics generated by the LDA model on the Twitter data set (Table \ref{tab:besthpsldatwitter}): Topic 0 encapsulates the environmental concerns associated with the circular economy, with a strong focus on `waste', `recycle', and `plastic'. The inclusion of `eu', `breakfreefromplastic', and `wastetrade' within this topic suggests a geographically and politically contextualised discussion. In Topic 1, terms like `need', `waste', and `package' alongside `sustainablepackage' indicate an exploration of how businesses can address sustainability through innovative packaging solutions.

In conclusion, BERTopic provided a deep extraction of CE topics through a language model (BERT), CorEx provided a unique separation of topics (highest coherence score), and LDA captured the underlying concepts of CE on Twitter, all of which provide valuable insight into our understanding of the public attention on the CE. In the next three subsections, we will visualise and discuss the topics generated by the best models on each dataset in more depth.

\begin{table}
\centering
\caption{The topic terms over 2 topics obtained from the LDA model with the best hyperparameter values on the Twitter dataset.}
\resizebox{\textwidth}{!}{
\begin{tabular}{c|l}
\bottomrule
\multicolumn{1}{c}{Topic} & \multicolumn{1}{c}{Topic Term}                                                                                       \\ 
\hline
0                         & circulareconomy, waste, recycle, plastic, sustainability, new, eu, breakfreefromplastic, wastetrade, wasteshipment   \\
1                         & circulareconomy, sustainability, circular, economy, sustainable, business, need, waste, package, sustainablepackage  \\
\bottomrule
\end{tabular}}
\label{tab:besthpsldatwitter}
\end{table}

\paragraph{Visualisation and Analysis} \label{sec:thepublic}

\begin{figure}
    \centering
    \includegraphics[scale=0.5]{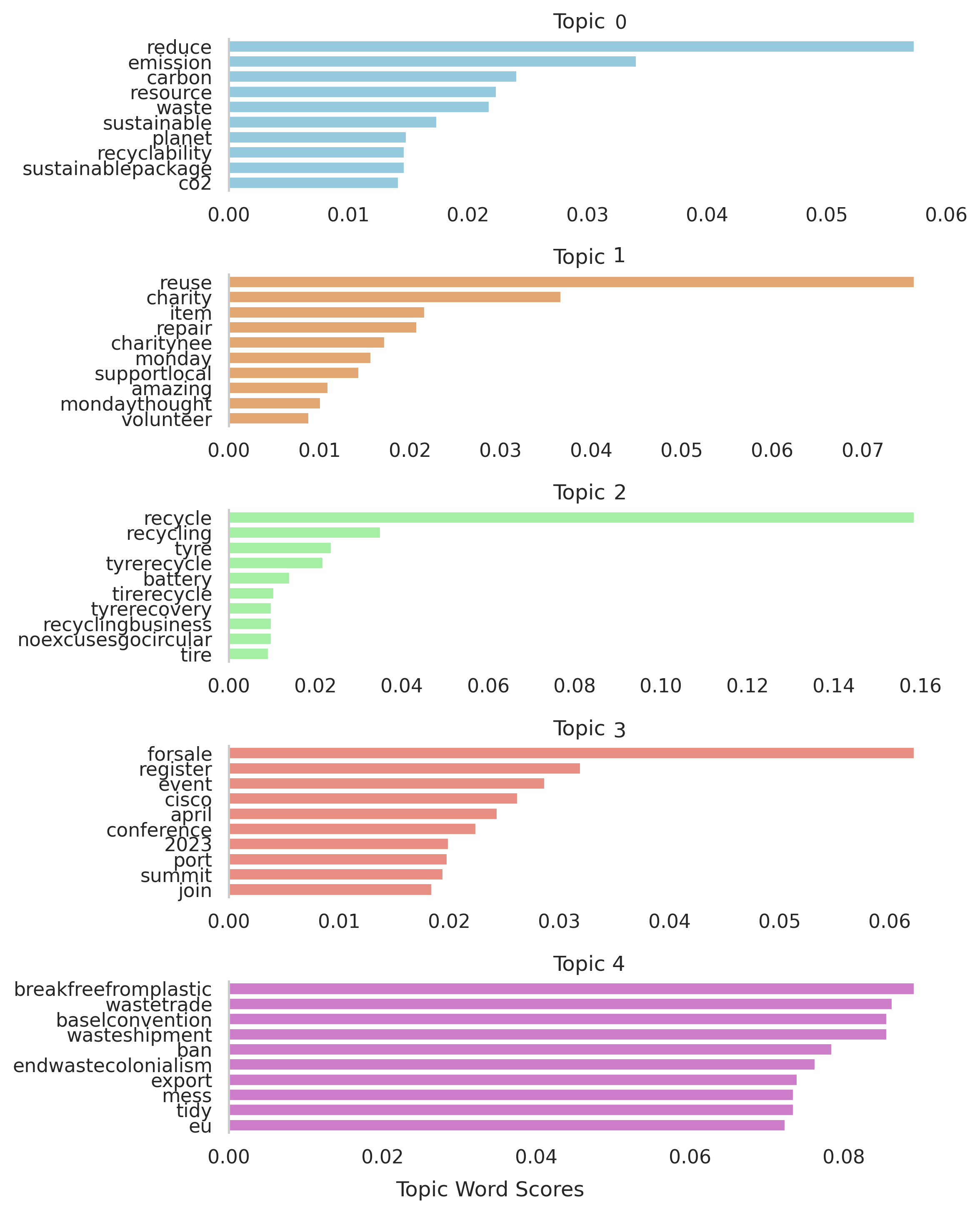}
    \caption{Bar chart representation of topic modelling results using the CorEx from Twitter dataset.}
    \label{fig:bar_twitter}
\end{figure}

This section analyses public participation and perspectives on the circular economy as reflected in the Twitter dataset. Using the CorEx topic model, Figure \ref{fig:bar_twitter} illustrates public attention and discussion on various aspects of the circular economy. The bar chart distinctly identifies the primary topics, revealing high concern for topics such as sustainability, recycling, and environment. This reflects significant public discourse surrounding sustainability and the circular economy. Specifically, Topic 0 focuses on sustainable practices and awareness, highlighting public attention to sustainable development and emphasising the need for ongoing ecological balance and responsible environmental management. Notably, terms like `emission', `reduce' frequently appear in discussions about the circular economy. The prominence of this term indicates public concern and interest in a low-carbon transition, likely spurred by growing awareness of environmental issues and the impacts of global climate change. Topic 2 centres on recyclability and waste management, showcasing public focus on the processes and benefits of recycling, and emphasising its crucial role in resource conservation and waste reduction. `recyclability' is a pivotal public concern, reflecting the importance of waste management in public discourse. The term’s high frequency can be attributed to the global movement and local regulations that promote recycling as a key component of resource conservation and the circular economy.

\begin{figure}
    \centering
    \includegraphics[scale=0.4]{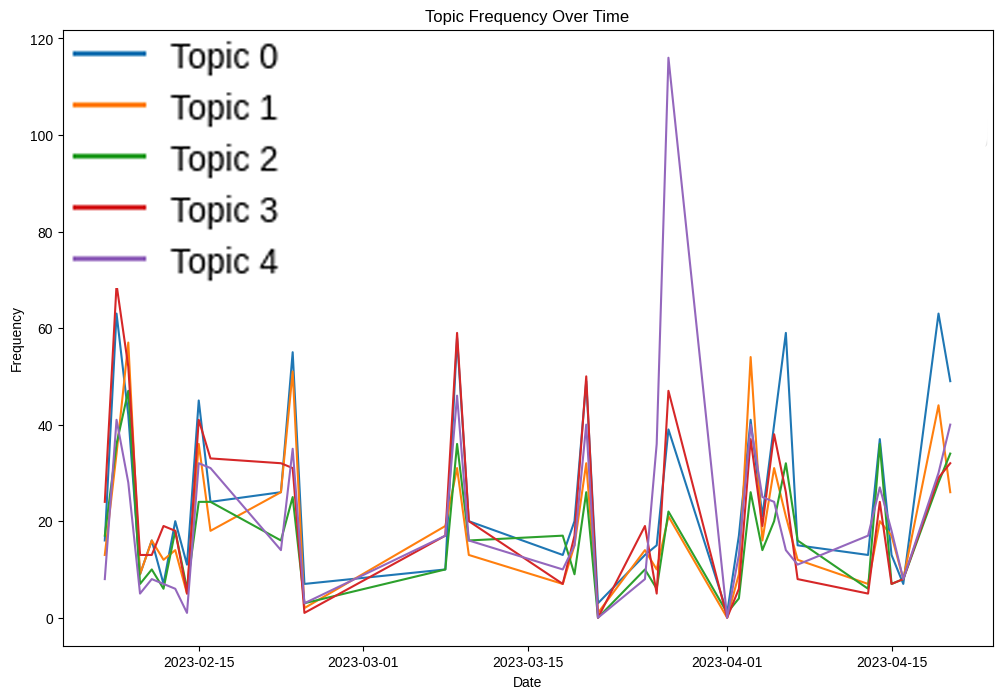}
    \caption{This line chart shows the frequency of occurrence of Topics generated by the CorEx model trained from the Twitter dataset in 2023.}
    \label{fig:dynamic_twitter}
\end{figure}

The line graph (Figure \ref{fig:dynamic_twitter}) displays the fluctuation frequencies of five distinct topics generated by the CorEx model based on 2023 Twitter dataset. These topics exhibit significant frequency peaks over time, which may correspond to specific events that triggered Twitter discussions or heightened public attention. Topic 2 shows less volatility compared to others but experiences a sharp increase in discussion volume. This likely relates to an event connected with `recycle', temporarily bringing this topic to the forefront on Twitter. Notably, Topic 4 demonstrates a very pronounced peak, exceeding the activity of other topics during the same period. This outlier could represent a major event or breakthrough in sustainability towards the end of March 2023, or an activity that briefly dominated discussions. Figure \ref{fig:dynamic_twitter} reflects Twitter’s reactive nature as a platform for public discourse. A temporal analysis of topic frequencies can provide deeper insights into how certain topics gain traction within the social media community, highlighting the interplay between global events and public attention.

\begin{figure}
    \centering
    \includegraphics[scale=0.4]{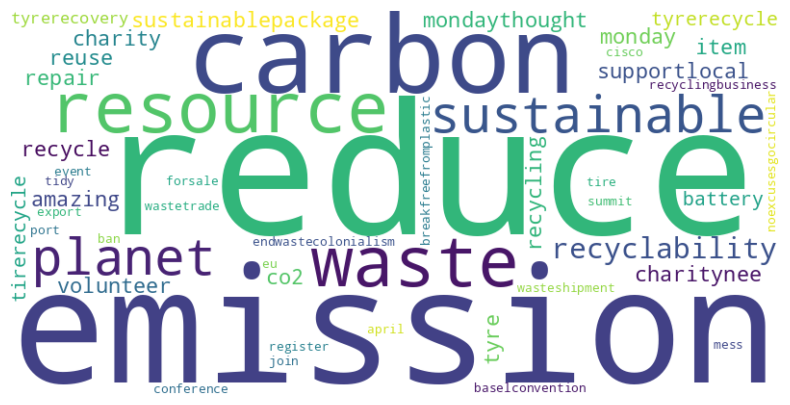}
    \caption{Word cloud visualisation of the prominence of terms in topics derived from Twitter dataset using the CorEx model.}
    \label{fig:wc_twitter}
\end{figure}

Further insights are provided by Figure \ref{fig:wc_twitter}, which displays a word cloud visualisation that highlights the frequency and significance of specific terms in public attention. The larger, more prominent words like `reduce', `carbon', `waste', and `emission' reflect a strong focus on environmental impact reduction and the urgency of addressing climate change through tangible actions such as waste management and carbon footprint minimisation. The term `sustainable', adjacent to `resource' and `planet', signifies the overarching goal of these discussions—maintaining the earth's ecological balance. Additionally, `recycle' and `recyclability' underscore the practical measures being discussed, likely linked to the optimisation of material usage and the lifestyle of products. Through this word cloud (\ref{fig:wc_twitter}), we can discern multiple aspects of the CE on Twitter. The topics generated by the CorEx model range from individual to broader societal behaviours, all of which assist researchers in playing a role in shaping a sustainable future within the circular economy.

\subsubsection{MOO} \label{sec:twitter_moo}

The topics generated by BERTopic on the Twitter dataset, optimised using MOO as shown in \Cref{tab:moo_topics_twitter}. For instance, Topic 0 includes terms like `circulareconomy', `recycling' and `circular', directly addressing key aspects of resource efficiency and waste management central to CE. The frequency of these terms indicates that users on Twitter are actively engaging with discussions about recycling and resource reuse, reflecting the platform's role in disseminating information and promoting environmental advocacy.

Topic 2 delves into the specifics of sustainable packaging and recycling, with terms such as `sustainablepackage' and `recycledmaterial' demonstrating how MOO can highlight niche topics within broader CE discussions. This kind of granular analysis is particularly useful for identifying targeted conversations about sustainability in industry, especially in relation to packaging materials.

Table \ref{tab:moo_topics_twitter} shows the diversity of topics, ranging from charity-related support initiatives (Topic 3) to political and legislative discussions on plastic waste (Topic 4). The ability of MOO to uncover such a wide array of topics underscores its value in capturing the breadth of public discourse on CE. Overall, MOO allows for a more detailed extraction of the evolving narratives on platforms like Twitter, revealing both grassroots actions and industry-specific concerns.

On the Twitter platform, the topic model generated by SOO optimisation exhibits relatively good consistency, especially with a smaller number of topics (e.g., 10), which focus on key subjects such as `emission', `carbon', and `recycling'. However, because SOO only optimises coherence, the generated topics lack diversity and fail to fully reflect the range of user discussions on Twitter, as well as the platform’s characteristic short text format.

MOO increases the number of topics (8) and improves the diversity score, capturing more discussions related to the circular economy. Although MOO’s consistency is slightly lower than that of SOO, the significantly reduced perplexity suggests that MOO better handles content on short-text platforms like Twitter. Additionally, MOO generates topics that cover a wider variety of subjects, such as electric vehicles (EVs) and environmental policies, making it more adaptable to the content diversity and brevity typical of Twitter. Therefore, MOO is better suited for platforms like Twitter, where topics are diverse and discussions are more dispersed.

\begin{table}
\centering
\caption{The topic terms obtained from the BERTopic model with the best hyperparameter values on the Twitter dataset (MOO).}
\resizebox{\textwidth}{!}{
\begin{tabular}{c|l}
\bottomrule
\multicolumn{1}{c}{Topic} & \multicolumn{1}{c}{Topic Term} \\ \hline
0  & circulareconomy, recycling, recycle, circular, sustainable, wastemanagement, sustainability, circularity, join, 2023 \\ 
1  & circulareconomy, circularity, circular, sustainabledevelopment, economic, future, economy, sustainable, consumption, reuse \\ 
2  & sustainablepackage, recycledmaterial, recyclability, sustainable, sustainability, recycling, recyclable, plasticpollution, flexiblepackage, recyclingbusiness \\ 
3  & charitynee, support, charityneed, supportlocal, recycle, charity, need, goodwill, goodwillfind, donate \\ 
4  & plasticwaste, wastetrade, endwastecolonialism, wasteshipment, waste, plasticpollution, recycling, eu, circulareconomy, ban \\ 
5  & c3650, cisco, gigabit, c6524gs, dl360, ethernet, port, 3650, server, uplink \\ 
6  & cisco1921, cisco, router, processor, performace, forsale, ethernet, 12000, c887vam, circulareconomy \\ 
7  & x450e, x450, x450a, x460, extreme, summit48si, poe, 24p, forsale, 24pt \\ 
\bottomrule
\end{tabular}}
\label{tab:moo_topics_twitter}
\end{table}

%% file: section5_clean.tex
\section{Conclusion, Policy Implications and Future Research Avenue}

\subsection{Conclusion}\label{sec:Conclusion}

In the context of sustainable transformation, this research aims to uncover public attention to the CE by analysing data from various sources, including social media platforms Twitter and Reddit, as well as the Guardian. By utilising topic modelling, this paper examines the multi-perspective and changing perceptions of the CE among the public, professionals, and officials. Three topic models (LDA, CorEx, and BERTopic) were employed. Our findings reveal a growing public concern for environmental issues, with a strong focus on sustainability, recycling, and circular practices. Furthermore, the dynamic evolution of these topics over time indicates that public interest in the CE is not only increasing but also diversifying. This reflects a deepening engagement with the CE, influenced by global and regional environmental events. Through systematic experiments and twin hyperparameter optimisation of the models, we effectively captured shifts in public attention and opinions, thereby providing a comprehensive view of societal concerns regarding the CE. Overall, this study contributes to both academic and practical understandings of public perceptions of the CE, mapping the cognitive pathways and concerns across different societal segments, thus offering valuable insights for policymakers and businesses to adapt strategies and communications for a more sustainable future.

This research compares and analyses the results of SOO (single-objective optimisation) and MOO (multi-objective optimisation) on CE-related topics across different platforms (The Guardian, Reddit, Twitter), highlighting the differences between these optimisation methods in topic model generation. SOO generates topics with high semantic relevance by maximising topic coherence, but it lacks diversity and adaptability, especially on platforms with dispersed discussions (such as Reddit and Twitter), where the number of generated topics is small and the information coverage is limited. In contrast, MOO produces a more diverse set of topics by optimising coherence, diversity, and perplexity simultaneously, demonstrating stronger adaptability on platforms with more complex and short-form texts. MOO not only enhances the model’s ability to capture diverse discussion content but also significantly reduces perplexity while maintaining topic coherence, enabling it to handle unseen texts more robustly. In summary, MOO’s application across different platforms proves it to be more suitable for analysing CE topics, providing a more comprehensive understanding of public concerns and discussions.

\subsection{Policy Implications}\label{sec:Policy Implications}

Our study also offers important policy implications. First, the growing public attention and concern regarding sustainability, recycling, and the CE underscores the need for targeted education programs to enhance public knowledge of CE principles \cite{Kirchherr2019}. Such programs should emphasise the benefits of circular practices, illustrating how individuals actions, like recycling and choosing sustainable products, contribute to environmental protection \cite{Scalabrino2022}. Additionally, community engagement can be designed to involve local residents in CE initiatives, thereby fostering public awareness and responsibility for environmental stewardship. Tailoring these policies to different regions and demographic groups will help maximise their reach and impact.
    
Furthermore, the study highlights strong public support for businesses that adopt CE practices (e.g., emissions reduction, reduce waste, use recycled materials, developing innovative circular products) \cite{Smol2018}. Policymakers should consider creating incentives (e.g., tax exemptions, subsidies) for businesses that integrate CE principles into their operations \cite{Domenech2019}. Additionally, policies should encourage companies to take greater responsibility in advancing the CE, ensuring that businesses operate within a clear regulatory framework that promotes sustainable development.
    
Given the significant and growing public attention to recycling and waste management, policymakers should enhance existing regulations to establish more stringent waste management policies and improved recycling processes \cite{Boffardi2021}. Measures could include setting higher recycling targets for government authorities, imposing stricter controls on industrial waste, and providing incentives for companies that significantly reduce waste production. Expanding waste management infrastructure will also be crucial to meeting growing public demand for efficient recycling services \cite{Tansel2020}. Moreover, policies should support the development of new technologies that improve recycling efficiency and waste reprocessing capabilities, ultimately making CE practices more accessible and effective.

\subsection{Future Research Avenue}\label{sec:Future Research Avenue}

Future research will focus on conducting deeper explorations based on the BERTopic model, which has shown optimal performance in both datasets used in this study. This includes investigating multimodal approaches that incorporate textual data as well as visual and auditory content, offering a more comprehensive analysis of public attention. This might involve integrating Transformer-based architectures \cite{vaswani2017} with computer vision techniques to analyse images and videos related to the CE on social media, thereby providing a richer and more thorough understanding of public discourse. This integrated approach not only enhances the precision of attention and thematic detection but also captures a broader array of data points for attention analysis, including emotional expressions and visual symbols. Longitudinal and cross-platform studies are crucial for observing changes in the emotions of different user groups over time, thereby gaining deeper insights into the dynamic shifts in public engagement with sustainability issues.

\section*{Data Statement}
All data used in our experiments are from The Guardian, Reddit and Twitter, which are publicly available and can be accessed by APIs in \href{https://open-platform.theguardian.com/}{The Guardian Open Platform}, \href{https://www.reddit.com/dev/api/}{Reddit API}, and \href{https://developer.twitter.com/en/docs/twitter-api}{Twitter API}.

\section*{Acknowledgements}
This research is funded by Digital Circular Electrochemical Economy (DCEE) \href{https://gow.epsrc.ukri.org/NGBOViewGrant.aspx?GrantRef=EP/V042432/1}{EP/V042432/1}, and the UK Research and Innovation (UKRI) Interdisciplinary Centre for Circular Chemical Economy \href{https://gow.epsrc.ukri.org/NGBOViewGrant.aspx?GrantRef=EP/V011863/1}{EP/V011863/1} and \href{https://gow.epsrc.ukri.org/NGBOViewGrant.aspx?GrantRef=EP/V011863/2}{EP/V011863/2}.